\documentclass{article}
\usepackage{iclr2021_conference,times}

\usepackage{amsmath,amsfonts,bm}

\def\eqref#1{equation~\ref{#1}}

\def\1{\bm{1}}

\DeclareMathAlphabet{\mathsfit}{\encodingdefault}{\sfdefault}{m}{sl}
\SetMathAlphabet{\mathsfit}{bold}{\encodingdefault}{\sfdefault}{bx}{n}

\newcommand{\E}{\mathbb{E}}

\newcommand{\KL}{D_{\mathrm{KL}}}

\usepackage[utf8]{inputenc}         
\usepackage[T1]{fontenc}            
\usepackage{hyperref}               
\usepackage{url}                    
\usepackage{booktabs}               
\usepackage{amsfonts}               
\usepackage{amssymb}                
\usepackage{mathtools}              
\usepackage{nicefrac}               
\usepackage{microtype}              
\usepackage{xcolor}                 
\usepackage{import}                 
\usepackage{amsmath}                
\usepackage{graphicx}               
\usepackage{tabularx}               
\usepackage{pgf, tikz}              
\usepackage[shortlabels]{enumitem}  
\usepackage{natbib}                 
\usepackage{multicol}               
\usepackage{wrapfig}                
\usepackage{subcaption}             
\usepackage{fancyhdr}               
\usepackage{lastpage}               
\usepackage{epigraph}               
\usepackage{scrextend}              
\usepackage{multicol}               
\usepackage{wrapfig}                
\usepackage{subcaption}             
\usepackage{scrextend}              
\usepackage{multirow}               
\usepackage{dblfloatfix}            
\usepackage{array}

\usepackage{booktabs}

\usepackage{custom_configs}         

\title{Self-Supervised Variational Auto-Encoders}

\author{Ioannis Gatopoulos \\
Department of Computer Science\\
University of Amsterdam / BrainCreators\\
Amsterdam, The Netherlands\\
\texttt{johngatop@gmail.com} \\
\And
Jakub M. Tomczak \\
Department of Computer Science \\
Vrije Universiteit Amsterdam \\
Amsterdam, The Netherlands \\
\texttt{jmk.tomczak@gmail.com}
}

\iclrfinalcopy 

\begin{document}

    \maketitle
    
    \begin{abstract}
    Density estimation, compression and data generation are crucial tasks in artificial intelligence. Variational Auto-Encoders (VAEs) constitute a single framework to achieve these goals.
Here, we present a novel class of generative models, called \textit{self-supervised Variational Auto-Encoder} (selfVAE), that utilizes deterministic and discrete variational posteriors.
This class of models allows to perform both conditional and unconditional sampling, while simplifying the objective function.
First, we use a single self-supervised transformation as a latent variable, where a transformation is either downscaling or edge detection.
Next, we consider a hierarchical architecture, i.e., multiple transformations, and we show its benefits compared to the VAE.
The flexibility of selfVAE in data reconstruction finds a particularly interesting use case in data compression tasks, where we can trade-off memory for better data quality, and vice-versa.
We present performance of our approach on three benchmark image data (Cifar10, Imagenette64, and CelebA).
\vskip -2mm


\vskip -3mm


    \end{abstract}
    
    \section{Introduction}
    \label{sec:introduction}
\begin{figure}[!b]
\centering
\scalebox{0.9}{
\begin{subfigure}{1.\textwidth}
  \centering
  \includegraphics[width=0.8\linewidth]{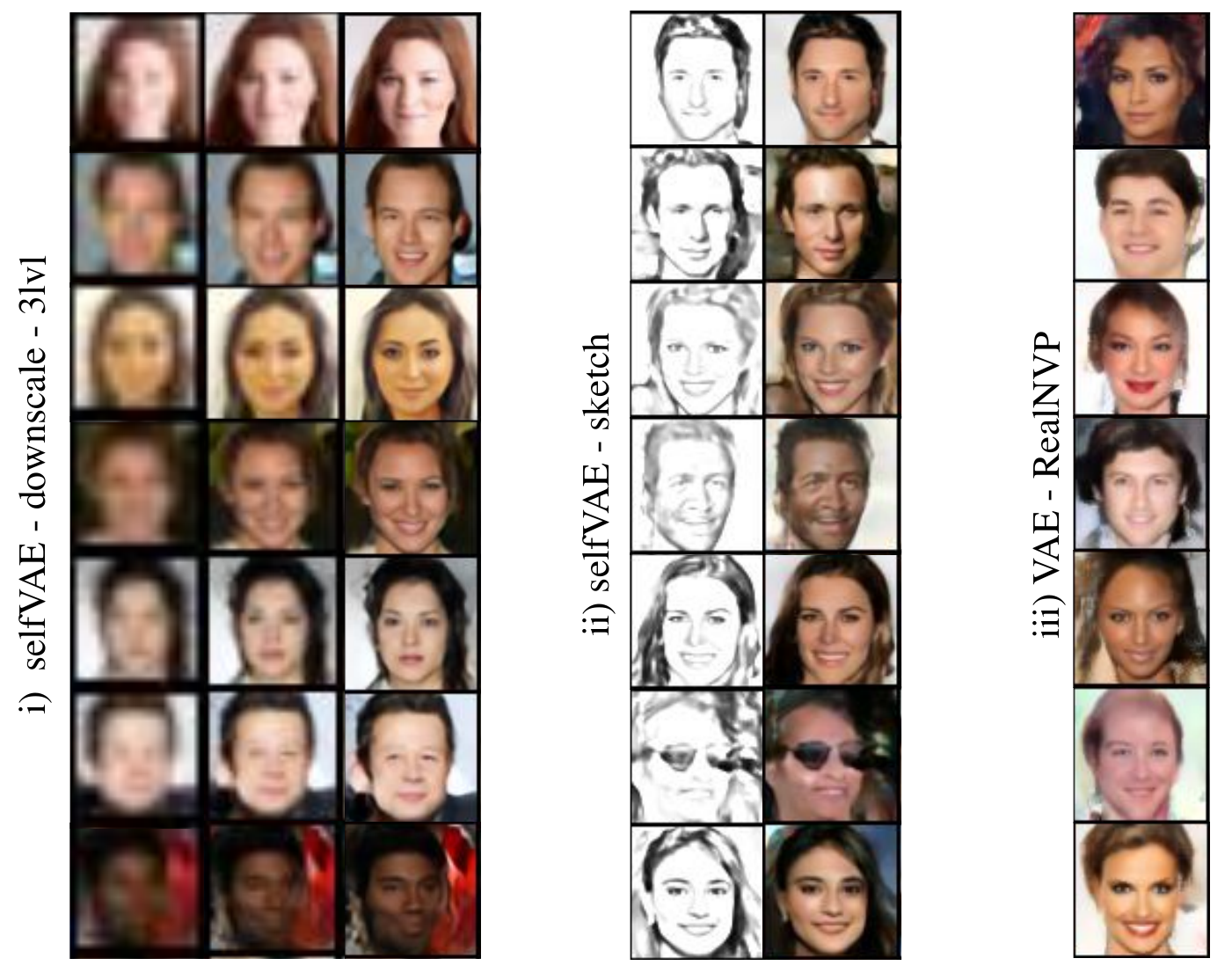}
\end{subfigure}}
\vskip -2mm
\caption{Unconditional CelebA generations from (i) the three-level self-supervised VAE with downscaling, (ii) the self-supervised VAE employed with edge detection (sketches), (iii) the VAE with RealNVP prior.}
\label{fig:celeba_generations_intro}
\end{figure}

The framework of variational autoencoders (VAEs) provides a principled method for jointly learning latent-variable models and corresponding inference models.
As it utilizes a meaningful low-dimensional latent space with density estimation capabilities, it forms an attractive solution for generative modelling tasks.
However, as its performance in terms of test likelihood and quality of generated samples was far from the desired one, many modifications were proposed in order to improved its performance on high-dimensional data like natural images.
In general, one can obtain a tighter lower bound, and, thus, a more powerful and flexible model, by advancing over the following three elements: the \textit{encoder} \citep{rezende2014stochastic, berg2018sylvester, hoogeboom2020convolution, maale2016auxiliary}, the \textit{prior} (or \textit{marginal} over latents) \citep{chen2016variational, habibian2019video, lavda2020data, lin2020ladder, tomczak2017vae} and the \textit{decoder} \citep{gulrajani2016pixelvae}.
Nevertheless, recent studies have shown that by employing deep hierarchical architectures and by carefully designed the building blocks of the neural networks, VAEs can successful model large high-dimensional data and reach state-of-the-art test likelihoods \citep{zhao2017learning, maale2019biva, vahdat2020nvae}.


In this work, we present a novel class of generative models, called \textit{self-supervised Variational Auto-Encoders}, where we introduce additional variables to VAEs that result from discrete and deterministic transformations of the given data samples, and specifically images.
Since the transformations are deterministic, namely, non-trainable, and they provide a specific aspect of images (e.g., contextual information through detecting edges or downscaling), we refer to them as \textit{self-supervised representations}.
The introduction of the discrete and deterministic variables allows to train deep hierarchical models efficiently and steadily, by breaking down the discovery of a highly complex distribution into smaller, overall simpler ones.
In this way, the model allows to integrate the prior knowledge about the data, but still enables to synthesize unconditional samples.
Furthermore, the discrete and deterministic variables could be used to conditionally reconstruct data, which could be of great use in data compression and super-resolution tasks.

In summary, we make the following contributions: i) We propose an extension of the VAE framework by employing a bijective prior and by incorporating self-supervised representations of the data. ii) We analyze the impact of modelling natural images with different data transformations as self-supervised representations. iii) This new type of generative model, namely, \textit{self-supervised Variational Auto-Encoders}, which is able to perform both conditional and unconditional sampling, demonstrate improved quantitative performance in terms of density estimation and generative capabilities on image benchmarks.
    
    \section{Background}
    \label{sec:background.tex}

\subsection{Variational Auto-Encoders}

Let $\mathbf{X} = \{ \mathbf{x}_1, ..., \mathbf{x}_N\}$ with $\mathbf{x}_n \in \mathcal{X}^{\mathrm{D}}$ be independent and identically distributed observable variables, where $\mathcal{X}$ could be a discrete set of values (e.g., $\mathcal{X} = \{0, \ldots, 255\}$).
Given that the dimensionality $D$ of real-world data is typically large, thus, resulting in a complex, high-dimensional distributions, calculating the marginal likelihood $p(\mathbf{X}) = \prod_{n=1}^{N} p(\mathbf{x}_{n})$ becomes a challenging problem.
Moreover, a latent variable model with latent variables $\mathbf{z} \in \mathbb{R}^{\mathrm{M}}$, $p_{\vartheta}(\mathbf{x}) = \int p_{\vartheta}(\mathbf{x}, \mathbf{z}) \mathrm{d} \mathbf{z}$, where $\vartheta$ denotes parameters, provides a framework for learning data representations.
However, the optimization  of $p_{\vartheta}(\mathbf{x})$ through maximum likelihood estimation (MLE) becomes infeasible due to the intractability of the integration at hand. 
One possible way of overcoming this issue and obtaining a highly scalable framework is by introducing a variational family of distributions $\mathcal{Q}$.
In order to identify its (amortized) member $q_{\phi}(\mathbf{z}|\mathbf{x})$, we aim at minimizing the Kullback-Leibler divergence ($\mathrm{KL}$) to the \textit{real} posterior $p(\mathbf{z}|\mathbf{x})$. 
In consequence, we can derive a tractable objective function, namely, the \textit{evidence lower bound} (ELBO) \citep{Jordan1998}:
\begin{align}
    \log p_{\vartheta}(\mathbf{x})
    \geq
    \mathbb{E}_{q_{\phi}(\mathbf{z} | \mathbf{x})}
    \left[\log p_{\theta}(\mathbf{x} | \mathbf{z})
    - \log q_{\phi}(\mathbf{z} | \mathbf{x}) +
    \log p_{\lambda}(\mathbf{z})
    \right]
    \nonumber
    \equiv
    \mathcal{L}(\theta, \phi, \lambda)
    \nonumber,
\end{align}
where $q_{\phi}(\mathbf{z} | \mathbf{x})$ is the variational posterior (or the \textit{encoder}), $p_{\theta}(\mathbf{x} | \mathbf{z})$ is the likelihood function (or the \textit{decoder}) and $p_{\lambda}(\mathbf{z})$ is the \textit{prior} over the latent variables, parameterized by $\phi$, $\theta$ and $\lambda$ respectively. 
The optimization is done efficiently by computing the expectation by Monte Carlo integration while exploiting the \textit{reparameterization trick} in order to obtain an unbiased estimator of the gradients. 
This generative framework is known as \textit{Variational Auto-Encoder} (VAE) \citep{kingma2013autoencoding, rezende2014stochastic}.


\subsection{VAEs with Bijective Prior}
\label{sec:vae_realnvp_main}

Even though the lower-bound suggests that the prior plays a crucial role in improving the variational bounds, usually it is modelled by a fixed distribution (i.e., a standard multivariate Gaussian). 
While being relatively simple and computationally cheap, a fixed prior is known to result in over-regularized models that tend to ignore more of the latent dimensions \citep{burda2015importance, hoffman2016elbo, tomczak2017vae}. 
Moreover, even with powerful encoders, VAEs may still fail to match the variational posterior to a unit Gaussian prior \citep{rosca2018distribution}.

However, it is possible to obtain a rich, multi-modal prior distribution $p(\mathbf{z})$ by using a \textit{bijective} (or \textit{flow-based}) model \citep{dinh2016density}. 
Formally, given a latent code $\mathbf{z} \sim q_Z(\mathbf{z}| \mathbf{x})$, a base distribution $p_V(\mathbf{v})$ over latent variables $\mathbf{v} \in V$, and $f: V \xrightarrow{} Z$ consisting of a sequence of $L$ diffeomorphic transformations\footnote{That is, invertible and differentiable transformations.}, where $f_i(\mathbf{v}_{i-1}) = \mathbf{v}_{i}$, $\mathbf{v}_{0} = \mathbf{v}$ and $\mathbf{v}_{L} = \mathbf{z}$, the \textit{change of variable} can be used sequentially to express the distribution of $\mathbf{z}$ as a function of $\mathbf{v}$ as follows:
\begin{align}
    \log p_{Z}(\mathbf{z})
    =
    \log p_{V}(\mathbf{v}) - \sum_{i=1}^{L} \log \left| \frac{\partial f_i(\mathbf{v}_{i-1})}{\partial \mathbf{v}_{i-1}} \right| 
    \nonumber,
\end{align}
where $ \left|\frac{\partial f_i(\mathbf{v}_{i-1})}{\partial \mathbf{v}_{i-1}} \right|$ is the Jacobian-determinant of the $i^{th}$ transformation.

Thus, using the transformed prior we end up with the following training objective function:
\begin{align}
    \mathcal{L}\left(\theta, \phi, \lambda\right)
    =
    \mathbb{E}_{q_{\phi}(\mathbf{z} | \mathbf{x})}
    \Big[
    \log p_{\theta}(\mathbf{x} | \mathbf{z})
    - 
    \log q_{\phi}(\mathbf{z} | \mathbf{x}) \,+
    \log p_{V}(\mathbf{v}_{0})
    \, + \,
    \sum_{i=1}^{L}
    \log\left|\frac{\partial f^{-1}_i(\mathbf{v}_{i})}{\partial \mathbf{v}_{i}} \right| \Big].
    \nonumber
\end{align}
In this work, we utilize RealNVP \citep{dinh2016density} as the prior, however, any other flow-based model could be used \citep{kingma2018glow}. For the experiments and ablation study that shows the impact of the bijective prior on VAEs, please refer to the appendix (Section \ref{sec:vae_realnvp}).

    \section{Method}
    \label{sec:method.tex}
\subsection{Motivation}

\begin{wrapfigure}{r}{0.4\textwidth}
    \centering
    \includegraphics[width=.9\linewidth]{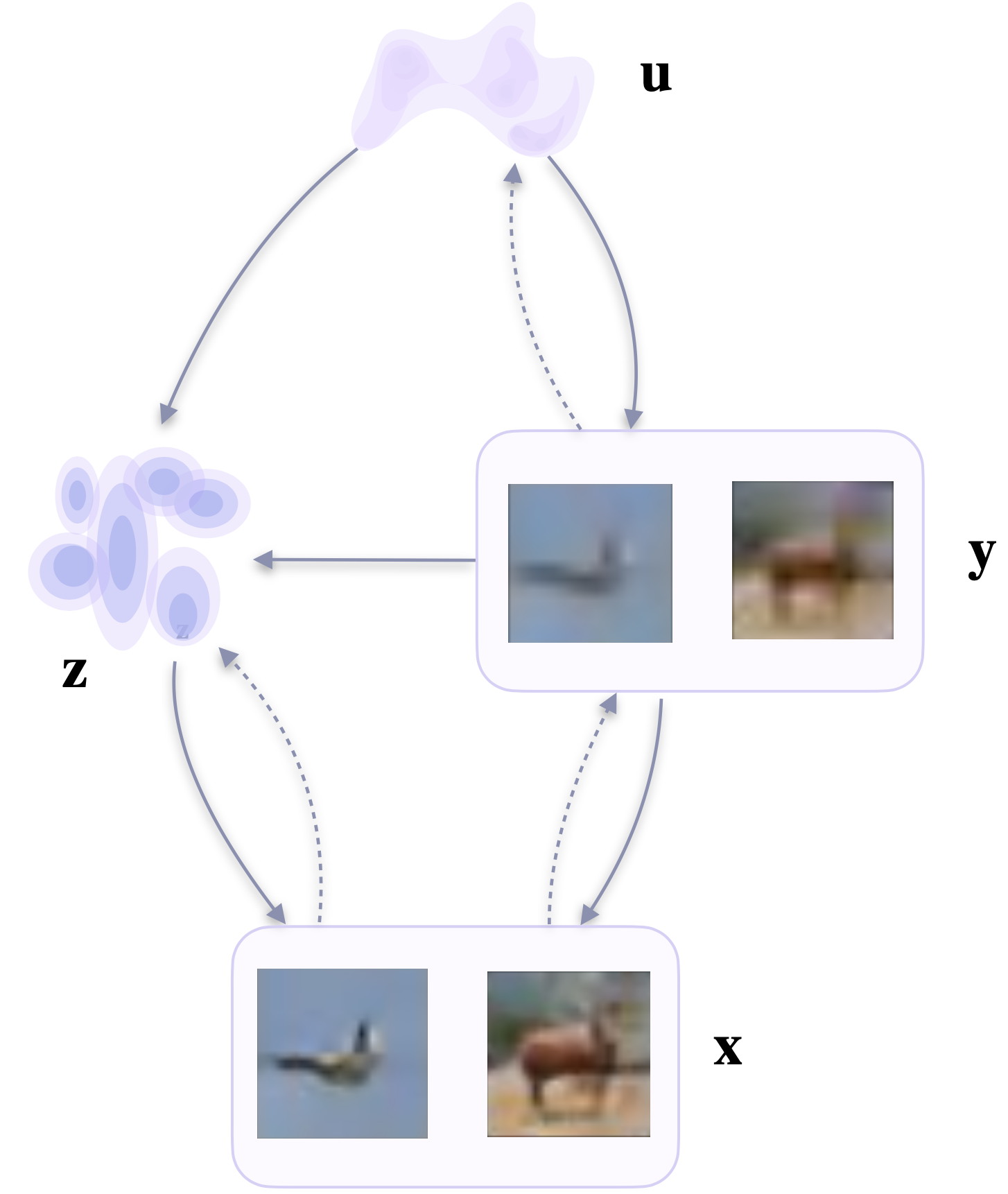}
    \vskip -2mm
    \caption{The proposed approach.}
    \label{fig:ssvae_graph_nice}
\end{wrapfigure}

The idea of self-supervised learning is about utilizing original unlabeled data to create additional context information.
It could be achieved in multiple manners, e.g., by adding noise to data \citep{vincent2008extracting} or masking data during training \citep{zhang2017split}.
Self-supervised learning could be also seen as turning an unsupervised model into a supervised one by, e.g., treating prediction of next pixels as a classification task \citep{henaff2019data, oord2018representation}.
These are only a few examples of a quickly growing research line \citep{liu2020self}.

Here, we propose to use non-trainable transformations to obtain information about image data.
Our main hypothesis is that since working with highly-quality images is difficult, we could alleviate this problem by additionally considering  partial information about them.
Thus, fitting a model to images of lower quality, and then enhancing them to match the target distribution seems to be overall an easier task \citep{chang2004super, gatopoulos2020superresolution}. By incorporating compressed transformations (i.e., the self-supervised representations) that still contain global information, with the premise that it would be easier to approximate, the process of modelling a high-dimensional complex density breaks down into  simpler tasks. 
In this way, the expressivity of the model will grow and gradually result into richer, better generations. 
A positive effect of the proposed framework is that the model allows us to integrate prior knowledge through the image transformations, without losing its unconditional generative functionality. 
Overall, we end up with a two-level VAE with three latent variables, where one is a data transformation that can be obtained in a self-supervised fashion. In Figure \ref{fig:ssvae_graph_nice} a schematic representation of the proposed approach with downscaling is presented.



A number of exemplary image transformations are presented in Figure \ref{fig:image_transformations}.
We can see that with these transformations, even though we discard a lot of information, the global structure is preserved. 
As a result, in practice the model should have the ability to extract a general concept of the data at the first stage, and add local information afterwards.
In this work, we focus on downscaling (Figure \ref{fig:image_transformations}.b, c \& d in the appendix) and edge detection or \textit{sketching} (Fig. \ref{fig:image_transformations}.i in the appendix). 



\subsection{Model formulation}

Let us introduce additional variables $\mathbf{y}\in \mathbb{R}^{\mathrm{C}}$ that result from self-supervised transformations of $\mathbf{x}$, where $C\leq D$. Further, let $\mathbf{u} \in \mathbb{R}^{N}$ and $\mathbf{z} \in \mathbb{R}^{M}$ be stochastic latent variables that interact with the above observed variables in a manner that is presented in Figure \ref{fig:ssvae_graph_nice}.
The dependencies of the considered model yield the following objective function (see the appendix \ref{sec:elbo_derivation} for the full derivation):
\begin{align}\label{eq:our_elbo}
    \mathcal{L}(\mathbf{x})
    =& \,
    \mathbb{E}_{q(\mathbf{z}| \mathbf{x}, \mathbf{y}) \, q(\mathbf{y}|\mathbf{x})} \log 
    p_{\theta}(\mathbf{x}| \mathbf{y}, \mathbf{z}) 
    - \E\mathord{_{q(\mathbf{u}|\mathbf{y})}} \KL({q(\mathbf{y}| \mathbf{x}})||p_{\theta}(\mathbf{y}| \mathbf{u}) ) +
    \nonumber
    \\&  
    -  \mathbb{E}_{q(\mathbf{u}|\mathbf{y}) q(\mathbf{y}|\mathbf{x})}\KL({q(\mathbf{z}| \mathbf{x}, \mathbf{y}) || p(\mathbf{z}| \mathbf{y}, \mathbf{u}})) 
    - \mathbb{E}_{q(\mathbf{y}|\mathbf{x})}\KL({q(\mathbf{u}| \mathbf{y}})||p(\mathbf{u})) ,
\end{align}
where $\KL(\cdot||\cdot)$ denotes the Kullback-Leibler divergence.

In addition, to simplify the training objective function and allow it to scale with high-dimensional data, we propose to consider the distribution $q(\mathbf{y}|\mathbf{x})$ as a \textit{deterministic} and \textit{discrete} density, i.e., a degenerate distribution (the Dirac's delta), that gives $\E\mathord{_{q(\mathbf{y}|\mathbf{x})}} \Big[ \log q(\mathbf{y}|\mathbf{x}) \Big] =0$. 
This property is important, because it does not introduce any complexity and allows us to exploit prior knowledge (i.e., image transformations) of the generative process, giving emphasis on specific desired characteristics of the processed data.
Next, we simplify the distribution $q(\mathbf{z}|\mathbf{y}, \mathbf{x})$ to $q(\mathbf{z}|\mathbf{x})$ since we assume that $\mathbf{y}$ does not introduce extra information about $\mathbf{x}$.
Thus, with these two properties in mind, the lower bound of the marginal likelihood of $\mathbf{x}$ is the following:
\begin{align}\label{eq:srvae_elbo}
    \mathcal{L}(\mathbf{x})
    =& \,
    \underbrace{\mathbb{E}_{q(\mathbf{z}| \mathbf{x}) \, q(\mathbf{y}|\mathbf{x})} \log 
    p_{\theta}(\mathbf{x}| \mathbf{y}, \mathbf{z})}_{\mathrm{RE}_{x}}
    + 
    \underbrace{\mathbb{E}_{q(\mathbf{u}|\mathbf{y}) q(\mathbf{y}|\mathbf{x})} \log p_{\theta}(\mathbf{y}| \mathbf{u})}_{\mathrm{RE}_{y}} +
    \nonumber
    \\&  
    - \underbrace{\mathbb{E}_{q(\mathbf{u}|\mathbf{y}) q(\mathbf{y}|\mathbf{x})} \KL({q(\mathbf{z}| \mathbf{x}) || p(\mathbf{z}| \mathbf{y}, \mathbf{u}}))}_{\mathrm{KL}_{z}}
    - \underbrace{\mathbb{E}_{q(\mathbf{y}|\mathbf{x})}\KL({q(\mathbf{u}| \mathbf{y}})||p(\mathbf{u}))}_{\mathrm{KL}_{u}} .
\end{align}

Further, in order to fully define our model, we propose to choose the following distributions:
\noindent
\begin{minipage}[t]{.5\textwidth}
    \vspace{-8pt}
     \begin{align}
        p(\mathbf{v}) 
        &= 
        \mathcal{N}\left(\mathbf{v} | \mathbf{0}, \mathbf{1})\right. \notag 
        \\
        p_{\lambda}\left(\mathbf{u}\right.) 
        &= 
        p(\mathbf{v}) \, \prod_{i=1}^{F} \Big|\operatorname{det} \frac{\partial f_i(\mathbf{v}_{i-1})}{\partial \mathbf{v}_{i-1}} \Big| ^{-1} \notag 
        \\
        p_{\theta_1}\left(\mathbf{y} | \mathbf{u}\right.) 
        &=
        \sum_{i=1}^{I} \pi_{i}^{(\mathbf{u})} \mathrm{Dlogistic}\Big(\mu_{i}^{(\mathbf{u})}, s_{i}^{(\mathbf{u})}\Big) \notag 
        \\
        p_{\theta_2}\left(\mathbf{z} | \mathbf{y}, \mathbf{u}\right.) 
        &=
        \mathcal{N}\left(\mathbf{z} | \boldsymbol{\mu}_{\theta_2}(\mathbf{y}, \mathbf{u}), \operatorname{diag}\left(\boldsymbol{\sigma}_{\theta_2}(\mathbf{y}, \mathbf{u})\right.) \right.) \notag 
        \\
        p_{\theta_3}\left(\mathbf{x} | \mathbf{z}, \mathbf{y}\right.)
        &= 
        \sum_{i=1}^{I} \pi_{i}^{(\mathbf{z}, \mathbf{y})} \mathrm{Dlogistic}\Big(\mu_{i}^{(\mathbf{z}, \mathbf{y})}, s_{i}^{(\mathbf{z}, \mathbf{y})} \Big) \notag 
     \end{align}
\end{minipage}%
\begin{minipage}[t]{.5\textwidth}
    \begin{align}
        \notag
        \\
        \notag
        \\
        q(\mathbf{y}|\mathbf{x}) 
        &= 
        \delta(\mathbf{y} = d(\mathbf{x})) \notag 
        \\
        \notag
        \\
        q_{\phi_1}\left(\mathbf{u} | \mathbf{y}\right.) 
        &=
        \mathcal{N}\left(\mathbf{u} | \boldsymbol{\mu}_{\phi_1}(\mathbf{y}), \mathrm{diag}\left(\boldsymbol{\sigma}_{\phi_1}(\mathbf{y})\right)  \right.) \notag 
        \\
        \notag
        \\
        q_{\phi_2}\left(\mathbf{z} | \mathbf{x}\right.) 
        &= 
        \mathcal{N}\left(\mathbf{z} | \boldsymbol{\mu}_{\phi_2}(\mathbf{x}), \mathrm{diag}\left(\boldsymbol{\sigma}_{\phi_2}(\mathbf{x})\right)  \right.) \notag 
    \end{align}
\end{minipage}

where $\mathrm{Dlogistic}$ is defined as the discretized logistic distribution \citep{salimans2017pixelcnn}, $\delta(\cdot)$ is the Dirac's delta, and $d(\mathbf{x})$ denotes the deterministic transformation that returns discrete values. 
In order to highlight the self-supervised part in our model, we refer to it as the \textit{self-supervised Variational Auto-Encoder} (or selfVAE for short).

\newpage

It is noteworthy that considering $\mathrm{RE}_{y}$ and $\mathrm{KL}_{u}$ in Eq. \ref{eq:srvae_elbo} alone is equivalent to a standard VAE objective, but for the transformed data $\mathbf{y}$.
Intuitively, the premise for selfVAE is that the latents $\mathbf{u}$ will capture the global structure of the input data and the latents $\mathbf{z}$ will encode the missing information between $\mathbf{y}$ and $\mathbf{x}$, guiding the model to discover the distribution of the target observations.

\begin{figure}[t]
\centering
\scalebox{.99}{
\begin{subfigure}{1.0\textwidth}
  \centering
  \includegraphics[width=1.\linewidth]{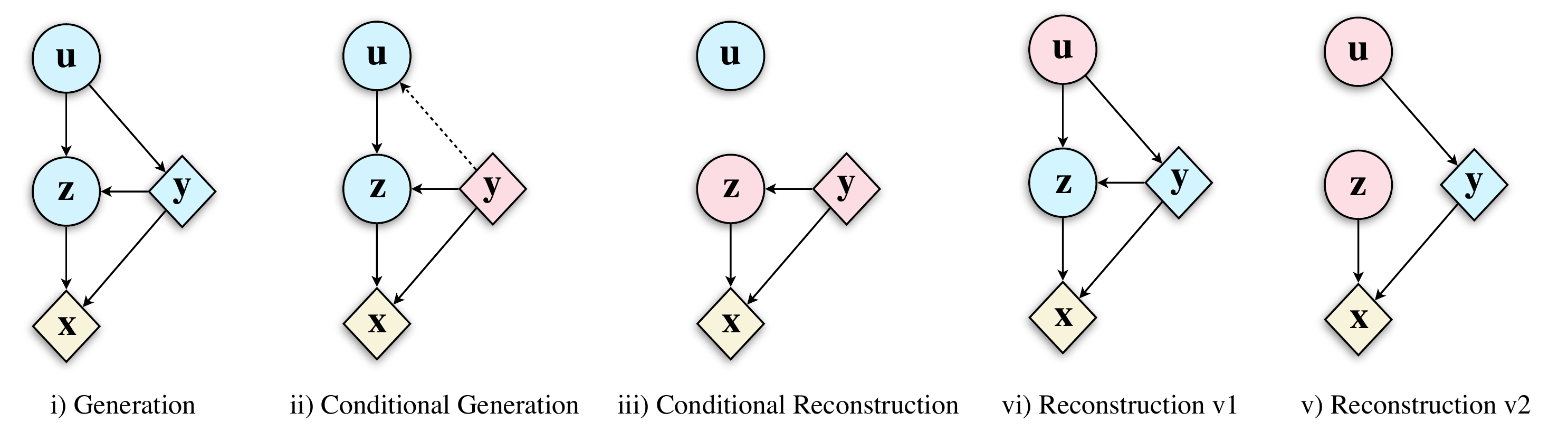}
\end{subfigure}}
\vskip -3mm
\caption{Operations of the self-supervised Variational Auto-Encoder. The blue and pink nodes represent the sampled and inferred latent codes, respectively. }
\label{fig:ssvae_reconstruction_graphs}
\end{figure} 

\subsection{Generation and Reconstruction in selfVAE}

As generative models, VAEs can be used to synthesize novel content through the following process:  $$\mathbf{z} \sim p(\mathbf{z}) \rightarrow \mathbf{x} \sim p(\mathbf{x}|\mathbf{z}),$$
but also to reconstruct a data sample $\mathbf{x}$ by using the following scheme: 
$$\mathbf{z} \sim q(\mathbf{z}|\mathbf{x}^{*}) \rightarrow \mathbf{x} \sim p(\mathbf{x}|\mathbf{z}).$$

Interestingly, our approach allows to utilize more operations regarding data generation and reconstruction, which are illustrated in Figure \ref{fig:ssvae_reconstruction_graphs}. First, analogously to VAEs, the selfVAE allows to generate data by applying the following hierarchical sampling process (\textit{generation}; Fig \ref{fig:ssvae_reconstruction_graphs}:i):  $$\mathbf{u} \sim p(\mathbf{u}) \xrightarrow{}
    \mathbf{y} \sim p(\mathbf{y}| \mathbf{u})
    \xrightarrow{}
    \mathbf{z} \sim p(\mathbf{z}| \mathbf{u}, \mathbf{y})
    \xrightarrow{}
    \mathbf{x} \sim p(\mathbf{x}| \mathbf{z}, \mathbf{y}).$$

Given that we can sample or infer the latent codes, we can use a variety of options on how to reconstruct or generate synthetic data $\mathbf{x}$. In particular, in \textit{conditional generation} and \textit{conditional reconstruction} (Fig \ref{fig:ssvae_reconstruction_graphs}:ii and iii) we use the ground-truth $\mathbf{y}$ (that is the $q(\mathbf{y}| \mathbf{x})$, a deterministic and discrete transformation) and sampled or inferred $\mathbf{z}$. The generative process for the former would be:
$$ \mathbf{z} \sim q(\mathbf{z}| \mathbf{x}^{*} )
    \xrightarrow{}
    \mathbf{x} \sim p(\mathbf{x}| \mathbf{z}, \mathbf{y}) ,$$
and for the latter:
$$\mathbf{u} \sim q(\mathbf{u}| \mathbf{y})
    \xrightarrow{}
    \mathbf{z} \sim p(\mathbf{z}| \mathbf{y}, \mathbf{u}),
    \xrightarrow{}
    \mathbf{x} \sim p(\mathbf{x}| \mathbf{z}, \mathbf{y}).$$

Notably, we can take advantage of the first process to tackle self-supervised or even supervised tasks, which relates to the chosen representation of $\mathbf{y}$. For example, if $\mathbf{y}$ is a downscaling transformation of the input image, it can be used in a manner similar to the super-resolution \cite{gatopoulos2020superresolution}.

Alternatively, given the flexibility of the model, we can instead sample (or generate) $\mathbf{y}$ and choose to sample or infer $\mathbf{z}$. In this way, we can reconstruct in two ways, namely, the \textit{reconstruction 1} (Figure \ref{fig:ssvae_reconstruction_graphs}:iv):
$$ \mathbf{y}^{*} \sim q(\mathbf{y}^{*}| \mathbf{x}^{*})
    \xrightarrow{}
    \mathbf{u} \sim q(\mathbf{u}| \mathbf{y}^{*})
    \xrightarrow{}
    \mathbf{y} \sim p(\mathbf{y}| \mathbf{u})
    \xrightarrow{}
    \mathbf{z} \sim p(\mathbf{z}| \mathbf{y}, \mathbf{u}),
    \xrightarrow{}
    \mathbf{x} \sim p(\mathbf{x}| \mathbf{z}, \mathbf{y}),
$$
and the \textit{reconstruction 2} (Figure \ref{fig:ssvae_reconstruction_graphs}:v):
$$ \Big{(} \mathbf{y}^{*} \sim q(\mathbf{y}^{*}| \mathbf{x}^{*} )
    \xrightarrow{}
    \mathbf{u} \sim q(\mathbf{u}| \mathbf{y}^{*})
    \xrightarrow{}
    \mathbf{y} \sim p(\mathbf{y}| \mathbf{u}) \Big{)}
    \text{ and }
    \mathbf{z} \sim q(\mathbf{z}| \mathbf{x}^{*} )
    \xrightarrow{}
    \mathbf{x} \sim p(\mathbf{x}| \mathbf{z}, \mathbf{y}).
$$
As we will see in the experiments, each option creates a different ratio of the reconstruction quality against the memory that we need to allocate to send information. Note, however, that every inferred variable needs to be sent, thus, more sampling corresponds to lower memory requirements.

\newpage
\subsection{Hierarchical self-supervised VAE}
\begin{wrapfigure}{r}{0.4\textwidth}
    \centering
    \includegraphics[width=.75\linewidth]{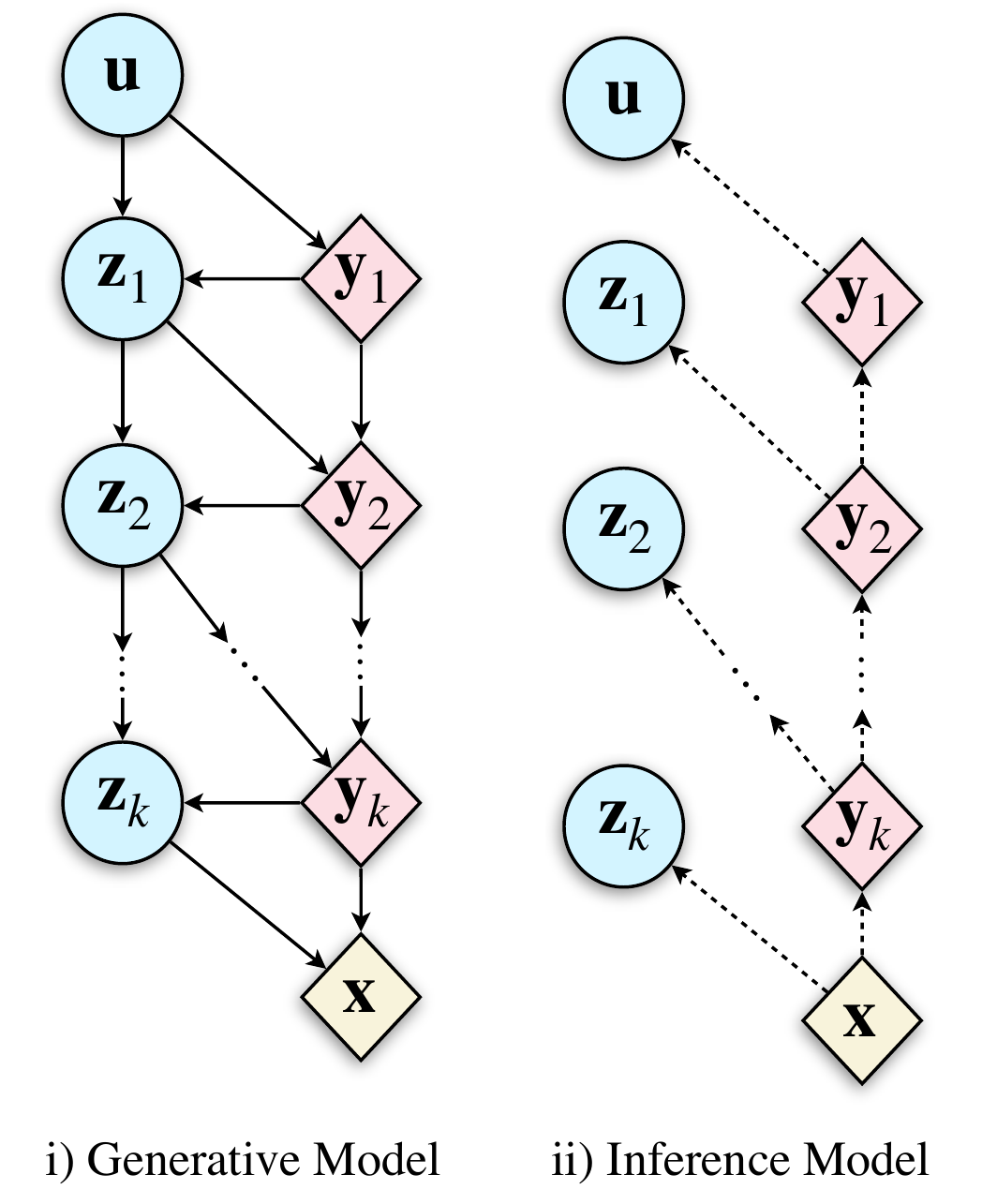}
    \vskip -3mm
    \caption{Hierarchical selfVAE.}
    \label{fig:ssvae_graph}
\end{wrapfigure}
The proposed approach can be further extended and generalized by introducing multiple transformations, in the way that it is illustrated in Figure \ref{fig:ssvae_graph}. 
By incorporating a single (or multiple) self-supervised representation(s) of the data, the process of modelling a high-dimensional complex density breaks down into $K$ simpler modeling tasks.
Thus, we obtain a $K$-level VAE architecture, where the overall expressivity of the model grows even further and gradually results into generations of higher quality. 
Some transformations cannot be applied multiple times (e.g., edge detection), however, others naturally could be used sequentially, e.g., downscaling.

Formally, the general case of Eq. (\ref{eq:srvae_elbo}) for $K$ self-supervised data transformations, where we define $\mathbf{z}_{0} = \mathbf{u}$ and $\mathbf{y}_{K+1} = \mathbf{x}$, can be expressed as follows:
\begin{align}\label{eq:ssvae_elbo}
    \mathcal{L}&(\mathbf{x})
    = \,
    \sum_{k=1}^{K}
    \mathbb{E}_{q(\mathbf{z}_{k}|\mathbf{y}_{k+1}) q(\mathbf{y}_{k}|\mathbf{y}_{k+1})} \log p_{\theta}(\mathbf{y}_{k+1}| \mathbf{y}_{k}, \mathbf{z}_{k})
    +
    \mathbb{E}_{q(\mathbf{u}|\mathbf{y}_{1}) q(\mathbf{y}_{1}|\mathbf{y}_{2})} \log p_{\theta}(\mathbf{y}_{1}| \mathbf{u}) +
    \nonumber
    \\& 
    - 
    \mathbb{E}_{q(\mathbf{y}_{1}|\mathbf{y}_{2})}\KL({q(\mathbf{u}| \mathbf{y}_{1}})||p(\mathbf{u})) 
    -
    \sum_{k=2}^{K} \mathbb{E}_{q(\mathbf{z}_{k-1}|\mathbf{y}_{k}) q(\mathbf{y}_{k}|\mathbf{y}_{k+1})} \KL({q(\mathbf{z}_{k}| \mathbf{y}_{k+1}) || p(\mathbf{z}_{k}| \mathbf{y}_{k}, \mathbf{z}_{k-1}})).
    \nonumber
\end{align}
    \section{Experiments}
    \label{sec:experiments}
\subsection{Experimental Setup}
We evaluate the proposed model on datasets of natural images: CIFAR-10, Imagenette64 and CelebA. We design a neural network architectures for the encoder and the decoder that consists building blocks composed of DenseNets \citep{huang2016densely} in order to encourage features reuse and help to preserve visual information, channel-wise attention \citep{zhang2018image} to favor the most informative channel features, and ELUs \citep{clevert2015fast} as activation functions.
The dimensionality of all the latent variables were kept at $8 \times 8 \times 16 = 1024$ and all models were trained using AdaMax optimizer \citep{kingma2014adam} with data-depended initialisation \citep{salimans2016weight}.
Regarding the selfVAEs, in CIFAR-10 we used an architecture with a single downscaled transformation (selfVAE-downscale), while on the remaining two datasets (CelebA and Imagenette64) we used a hierarchical 3-leveled selfVAE with downscaling, and a selfVAE with sketching. 
All models were employed with the bijective prior (RealNVP) comparable in terms of the number of parameters (the range of the weights of all models was from $32$M to $42$M). 
For more information about the neural network architecture and the datasets that where used, please refer to the appendix sections \ref{sec:nn} and \ref{sec:datasets}, respectively.
As evaluation metrics, we approximate the negative log-likelihood (\textit{nll}) using 512 IW-samples \citep{burda2015importance} and express the scores in bits per dimension (\textit{bpd}). 
Additionally, for CIFAR-10, we use the \textit{Fréchet Inception Distance} (FID) \citep{heusel2017gans}.

\vskip -4mm
\subsection{Quantitative results}

\begin{table*}[!b]
\caption{Quantitative comparison on test sets from CIFAR-10, CelebA, and Imagenette64. *Measured on training set.}
\centering
\scalebox{.85}{
\begin{tabular}{llcccc}
\textbf{Dataset} & \textbf{Model} & \textit{bpd} & RE & KL &
FID\\
 \hline
 \multirow{7}{*}{CIFAR-10} & PixelCNN \citep{oord2016pixel} & 3.14 & - & - & 65.93 \\
  & GLOW \cite{kingma2018glow} & 3.35 & - & - & 65.93 \\
 & ResidualFlow \citep{chen2019residual} & 3.28 & - & - & 46.37 \\
 & BIVA \citep{maale2019biva} & 3.08 & - & - & - \\
 & NVAE \cite{vahdat2020nvae} & \textbf{2.91} & - & - & - \\
 & WGAN-GP \citep{gulrajani2017improved} & - & - & - & 36.40 \\
 & VAE (ours) & 3.51 & 5540 & 1966 &  41.36 (37.25*) \\
 & selfVAE-downscale & 3.65 & 6348 & 1438 & \textbf{34.71} (\textbf{29.95}*) \\ \hline
\multirow{3}{*}{CelebA} & VAE (ours) & 3.12 & 24096 & 2502 & - \\
& selfVAE-sketch & 3.24 & 23797 & 3816 & - \\
& selfVAE-downscale-3lvl & \textbf{2.88} & 23336 & 1214 & - \\ \hline
\multirow{2}{*}{Imagenette64} & VAE (ours) & 3.85 & 30809 & 1961 & -  \\
 & selfVAE-downscale-3lvl & \textbf{3.70} & 30394 & 1171 & - \\
 \bottomrule
\end{tabular}}
\label{tab:results}
\end{table*}

\begin{figure}[t]
\centering
\scalebox{.99}{
\begin{subfigure}{1.\textwidth}
  \centering
  \includegraphics[width=1.\linewidth]{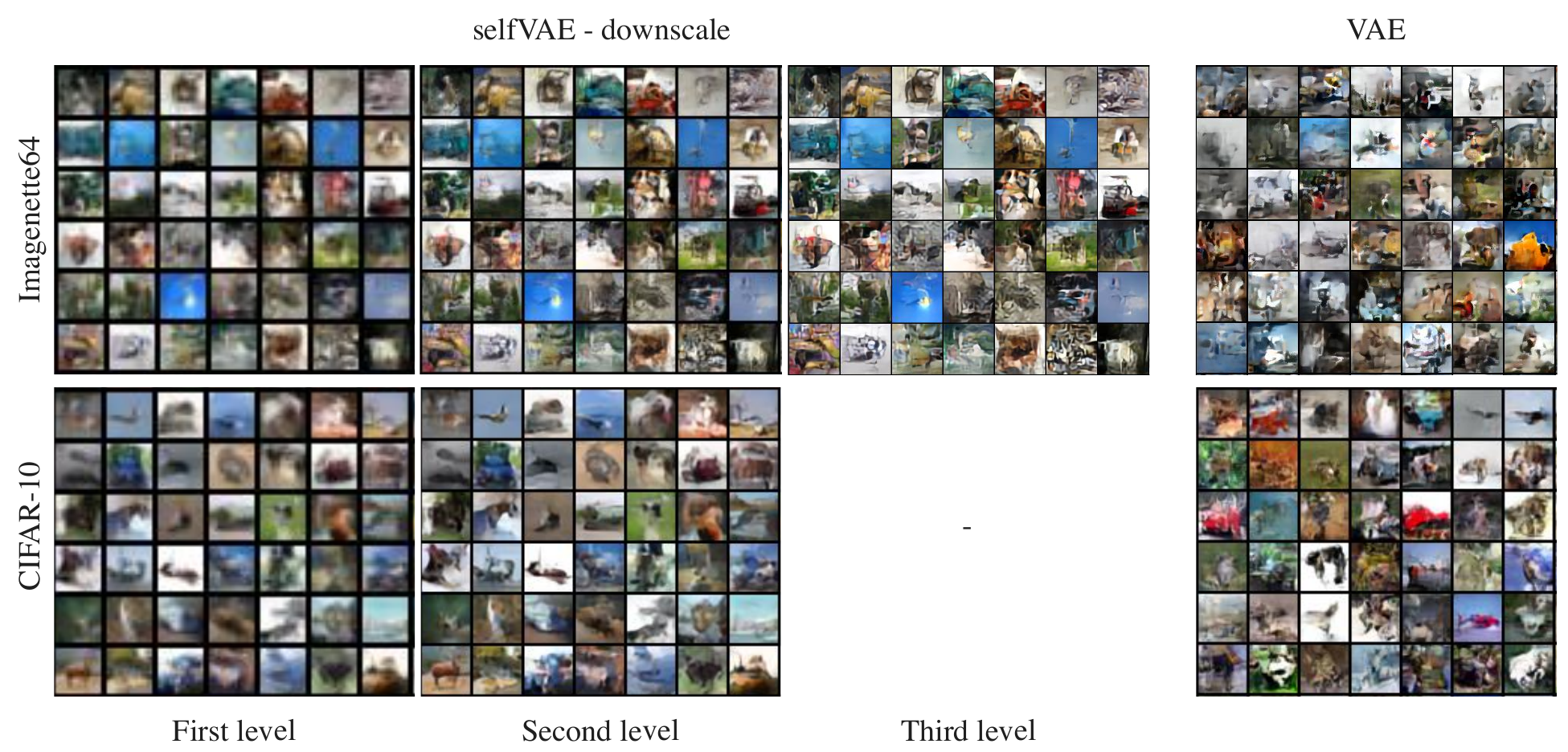}
\end{subfigure}}
\vskip -3mm
\caption{Uncoditional generations on Imagenette64 and CIFAR-10.}
\label{fig:gens}
\end{figure}

We present the results of the experiments on the three benchmark datasets in Table \ref{tab:results}. First we notice that on CIFAR-10 our implementation of the VAE is still lacking behind other generative models in terms of \textit{bpd}, however, it is better or comparable in terms of FID. The selfVAE-downscale achieves worse \textit{bpd} than the VAE. A possible explanation may lie in the small image size ($32$px $\times$ $32$px), as the benefits of breaking down the learning process in two or more steps are not obvious given the small target dimensional space.
Nevertheless, the selfVAE-downscale achieves significantly better FID scores than the other generative models. This result could follow from the fact that downscaling allows to maintain context information about the original image and, as a result, a general coherence is of higher quality.

Interestingly, on the two other datasets, a three-level selfVAE-downscale achieves significantly better \textit{bpd} scores than the VAE with the bijective prior. This indicates that the benefit of employing a multi-leveled self-supervised framework against the VAE is more obvious when we move to higher-dimensional data, where the plain model fails to scale efficiently. Unlike before, the RE is now better, but the biggest difference lies in the KL term. Apparently, it seems that the hierarchical structure of self-supervised random variables allows to encode the missing information more efficiently in $\mathbf{z}_k$, in contrast to the  vanilla VAE, where all information about images must be coded in $\mathbf{z}$. This result is promising and indicates that the proposed approach of using information in data in a self-supervised manner is of great potential for generative modelling.

\vskip -3mm
\subsection{Qualitative results}

\begin{figure}[t]
\centering
\scalebox{.98}{
\begin{subfigure}{1.\textwidth}
  \centering
  \includegraphics[width=1.\linewidth]{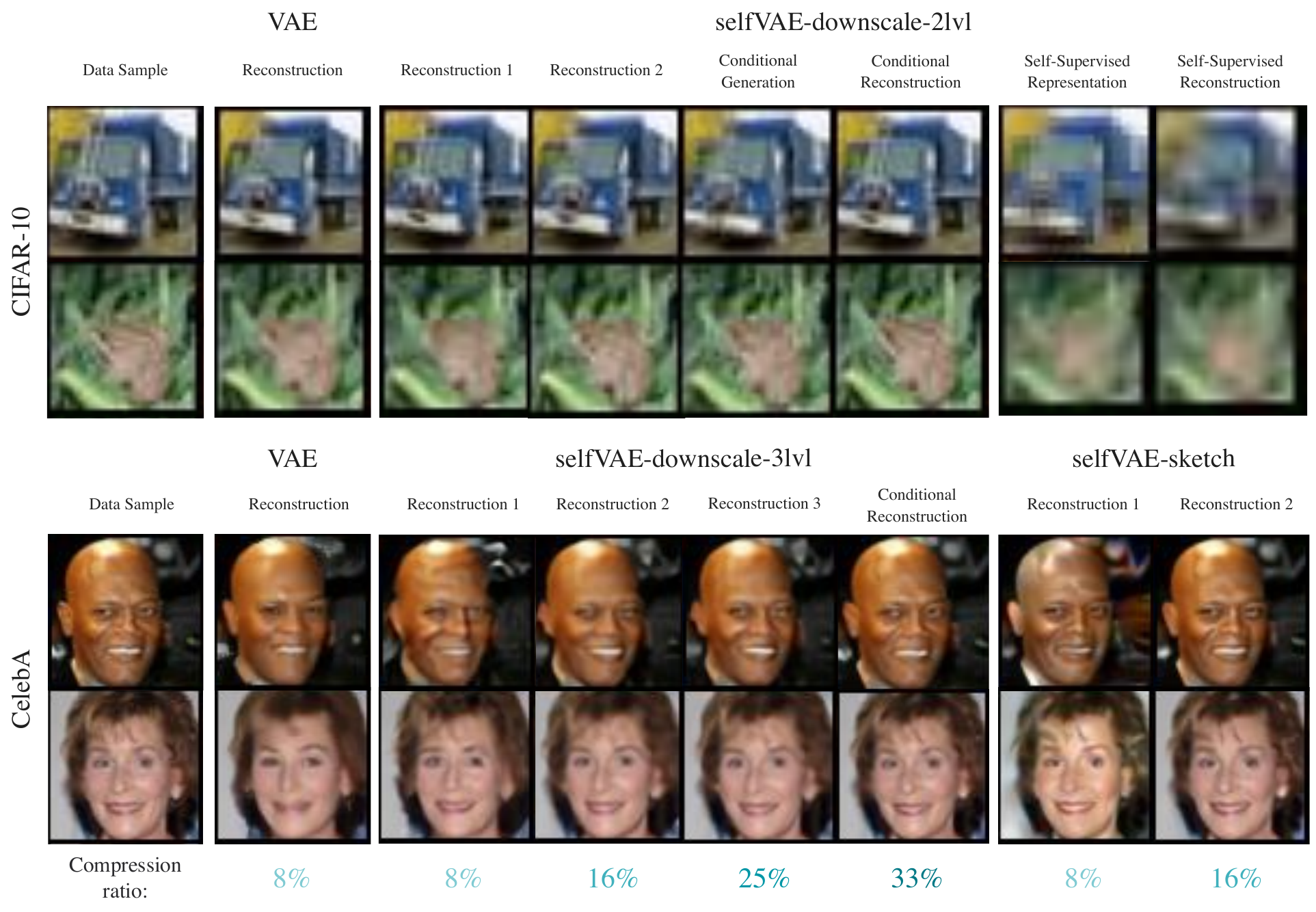}
\end{subfigure}}
\vskip -3mm
\caption{Comparison on image reconstructions with different amount of sent information.}
\label{fig:recon}
\vskip -4mm
\end{figure}

We present generations on CIFAR-10 and Imagenette64 in Figure \ref{fig:gens} and on CelebA in Figure \ref{fig:celeba_generations_intro}. 
In addition, we provide reconstructions on CIFAR-10 and CelebA in Figure \ref{fig:recon}. 
We first notice that the generations from selfVAE seem to be more coherent, in contrast with these from VAE that produces overall more contextless and distorted generations. 
This result seems to be in line with the FID scores. 
Especially for CelebA, we observe impressive synthesis quality, great sampling diversity and coherent generations (Figure \ref{fig:celeba_generations_intro}).
On the Imagenette64 dataset, we can also observe crisper generations for our method compared to the VAE (Figure \ref{fig:gens}).
Furthermore, the hierarchical selfVAE seems to be of a great potential for compression purposes. 
In contrast to the VAE, which is restricted to using a single way of reconstructing an image, the selfVAE allows four various options with different quality/memory ratios (see Figure \ref{fig:recon}).
Impressively, with the selfVAE-sketch we can retrieve the image with high accuracy by using only 16\% of the original data, as it manages to encode all the texture of the image to $\mathbf{z}$ (see Figure \ref{fig:recon_sketch} in the appendix). 
This shows the advantage of choosing prior knowledge into the learning process.
Lastly, the selfVAE-downscale can perform super-resolution out-of-the-box successfully that further highlights the importance of encoding details in $\mathbf{z}$. The latents $\mathbf{z}$ learns to add extra information, which defines the end result, and, when it sampled, we can alter details of an image like facial expressions (see Figure \ref{fig:latent}.ii in the appendix).

\vskip -1mm
    \section{Conclusion}
    \label{sec:section_5}
    \vskip -3mm

We proposed a novel class of generative models, called self-supervised Variational Auto-Encoder (selfVAE), where we extend the plain VAE architecture with self-supervised representations.
The application of transforming data in a self-supervised manner allows to integrate prior knowledge without loosing its unconditional generative functionality.
We showed that taking deterministic and discrete transformations results in a simplified ELBO which incorporates self-supervised representations. 
In the experimental studies, we indicated that the proposed approach results in coherent generations of high visual quality.
Moreover, the results seem to confirm that hierarchical architectures perform better and allow to obtain both better \textit{bpd} scores and better generations and reconstructions.
In this paper, we considered two classes of image transformations, namely, \textit{downscaling} and edge detection (\textit{sketching}). 
However, there is a vast of possible other transformations (see Figure \ref{fig:image_transformations}), and we leave investigating them for future work. 
Moreover, we find the proposed approach interesting for the compression task.
A similar approach with a multi-scale auto-encoder for image compression was proposed, e.g, by \cite{mentzer2019practical} or \cite{razavi2019generating}. However, we still use a probabilistic framework and indicate that various non-trainable image transformations (not only multiple scales) could be of great potential.

\vskip -5mm

    \section*{Acknowledgments}
    \label{sec:acknowledgments}
    \vskip -2mm
The authors would like to thank Maarten Stol (BrainCreators) and Efstratios Gavves (University of Amsterdam) for their support and fruitful discussions.


    \bibliography{main}
    \bibliographystyle{iclr2021_conference}
    
    \newpage
    \appendix
    \section{Appendix}

\subsection{Non-trainable image transformations}
\label{sec:image_transformations}
\begin{figure}[ht]
\centering
\scalebox{0.9}{
\begin{subfigure}{1.0\textwidth}
  \centering
  \includegraphics[width=1.\linewidth]{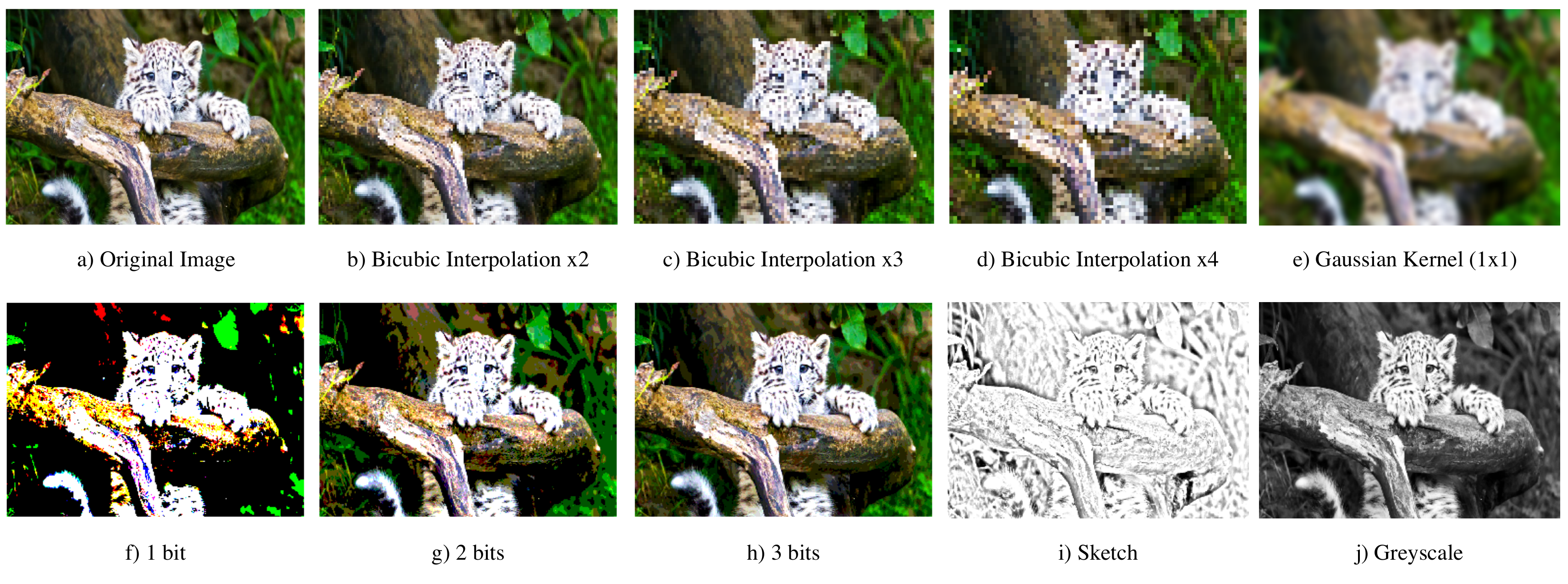}
\end{subfigure}}
\vskip -3mm
\caption[Image Transformations]{Image Transformations. All of these transformations still preserve the global structure of the samples but they disregard the high resolution details in different ways.}
\label{fig:image_transformations}
\end{figure}

\subsection{The impact of the bijective prior on VAEs}
\label{sec:vae_realnvp}

\subsubsection{Motivation}

\begin{figure}[h]
\centering
\scalebox{.5}{
\begin{subfigure}{1.\textwidth}
  \centering
  \includegraphics[width=1.\linewidth]{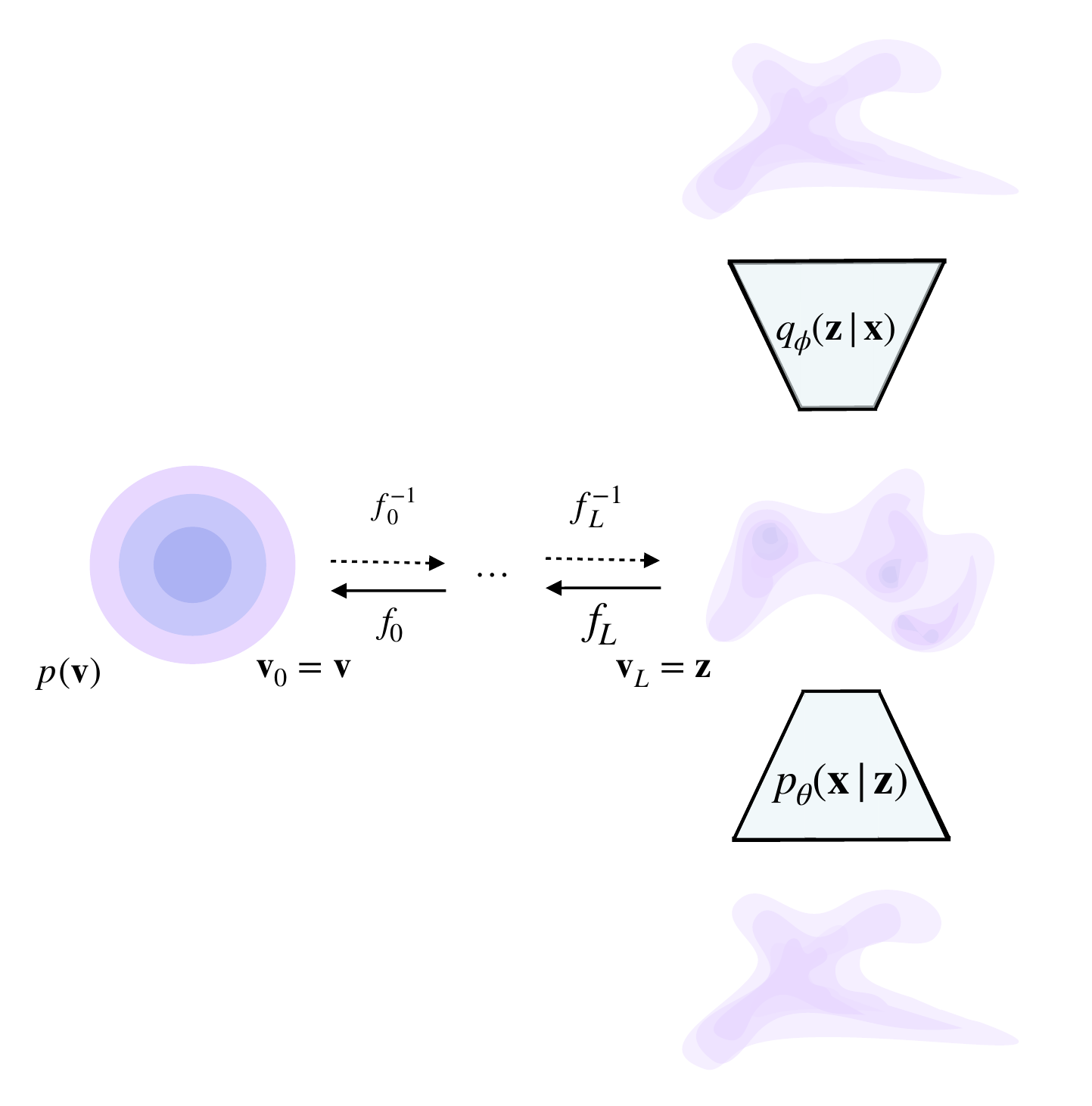}
\end{subfigure}}
\caption{VAE employed with bijective prior.}
\label{fig:vae_bijective}
\end{figure}

Bijective neural networks were first introduced to increase the expressibility of the encoder, due to its forced limitations to follow a mean-field variational family of Gaussian distributions. As they demand the base function to be known and simple (i.e. a distribution that is easy to estimate and sample from), the mapping of the input space to the latent follows a two step process; first the input sample is mapped to the base distribution though the encoder, and then transformed to match the unimodal Gaussian prior. However, despite the richer variational posterior, \cite{rosca2018distribution} proved that the optimal prior may still not much a fixed unit Gaussian distribution, which in addition, is known to result in over-regularized models that tend to ignore more of the latent codes \citep{burda2015importance, tomczak2017vae}.

However, little attention has been paid to adapt bijective neural networks as a learnable, data-driven prior of Variational Auto-Encoders. Since the prior of the latent dimensions plays a crucial part in the objective function, it will contribute into a better likelihood estimation of the data, while providing more informative latent codes. There are a couple of advantages using a bijective network to estimate the prior distribution:

\begin{itemize}
	\item[-] \textbf{Arbitrary complex, data-driven prior} \, Instead of forcing the encoder to match a fixed distribution, we follow the intuitive process, that now, the prior distribution adapts to the posterior during the training progress. Thus, the posterior is not being dragged to the centre of a standard Gaussian distribution and it is allowed to move freely in the latent space. As the normalizing flows framework allows the transformation of a simple function to an arbitrary complex one, the result would be a richer, multi-modal distribution that incorporates causal knowledge of the data at hand.

	\item[-] \textbf{Simple implementation and integration} \, Instead of breaking the encoder into a two step process, we fit a flow-based generative model to the latent codes produced by the encoder. Additionally, getting advantage of the invertible nature of the network, we can generate a datapoint by simply sampling a point from its base distribution, and transforming it (through the inverse process) into a latent code of the variational posterior.

	\item[-] \textbf{Fast sampling and inference}. An alternative way of retrieving a powerful latent prior is through autoregressive modelling of the latent codes. Formally, $p(\mathbf{z})$ is factorized as 
		\begin{align*}
			p(\mathbf{z})=\prod_{i=1}^{d} p\left(z_{i} | \mathbf{z}_{<i}\right)
		\end{align*}
		so that estimating $p(\mathbf{z})$ reduces to the estimation of each single conditional probability density (CPD) expressed as $p(z|\mathbf{z}_{<i})$, where the symbol $<$ implies an order over random variables. However, because high-dimensional data often requires a relatively high-dimensional latent space, in order to extract the highly informative causal components, the sampling process can be an important barrier for fast generation of novel content. On the contrary, the bijective neural networks require only a forward pass of a sampled point, reducing significantly the inference time.
\end{itemize}

It is straightforward to obtain a rich, multi-modal prior, and the process its described on section \ref{sec:vae_realnvp_main}.

\begin{figure}[t]
\centering
\scalebox{.7}{
\begin{subfigure}{1.\textwidth}
  \centering
  \includegraphics[width=1.\linewidth]{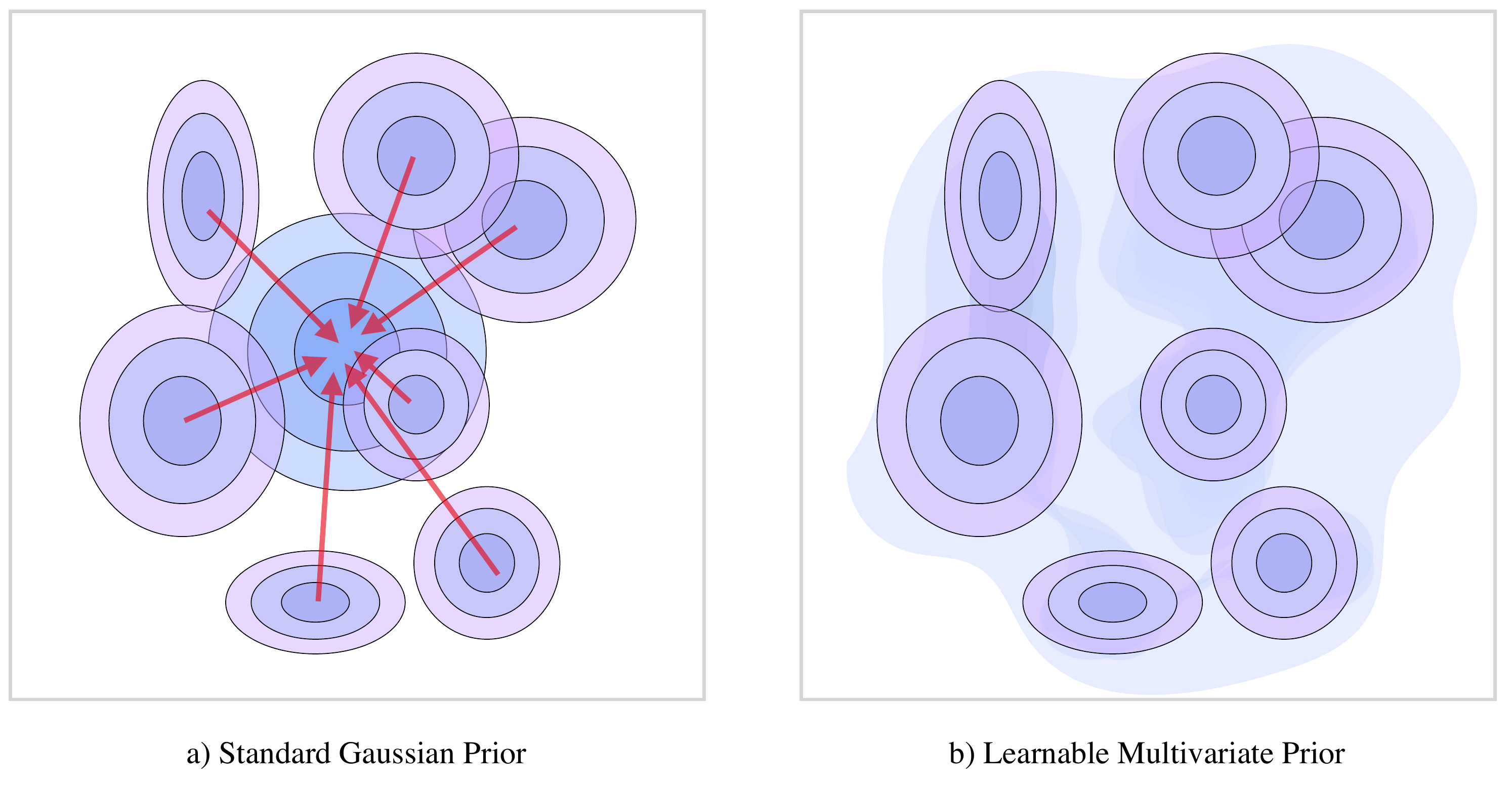}
\end{subfigure}}
\caption[Illustration of the Prior]{
Left: The standard prior is too strong and over-regularizes the encoder and create "holes" in the latent space. Right: Learnable prior adjusts to the posterior distribution, surrounding the variable posterior with smooth transitions.}
\label{fig:prior}
\end{figure}

\subsubsection{Experiments}

On this section, we will evaluate the performance of Variational Auto-Encoder with using different latent variable prior distributions, both on a quantitatively and qualitatively view. The models named as "VAE" represents the plain architecture with standard normal prior, while "VAE + MoG" and "VAE + RealNVP" represent data-driven priors, where the former employs $10$ mixture of Gaussians and the latter the bijective network RealNVP. For fair comparison, all the models share the same neural networks architectures both on the encoder and decoder. For the specific architecture of the neural networks, please refer to the section \label{sec:nn}.

\begin{figure}[t]
    \centering
    \scalebox{.99}{
    \begin{subfigure}{1.\textwidth}
      \centering
      \includegraphics[width=1.\linewidth]{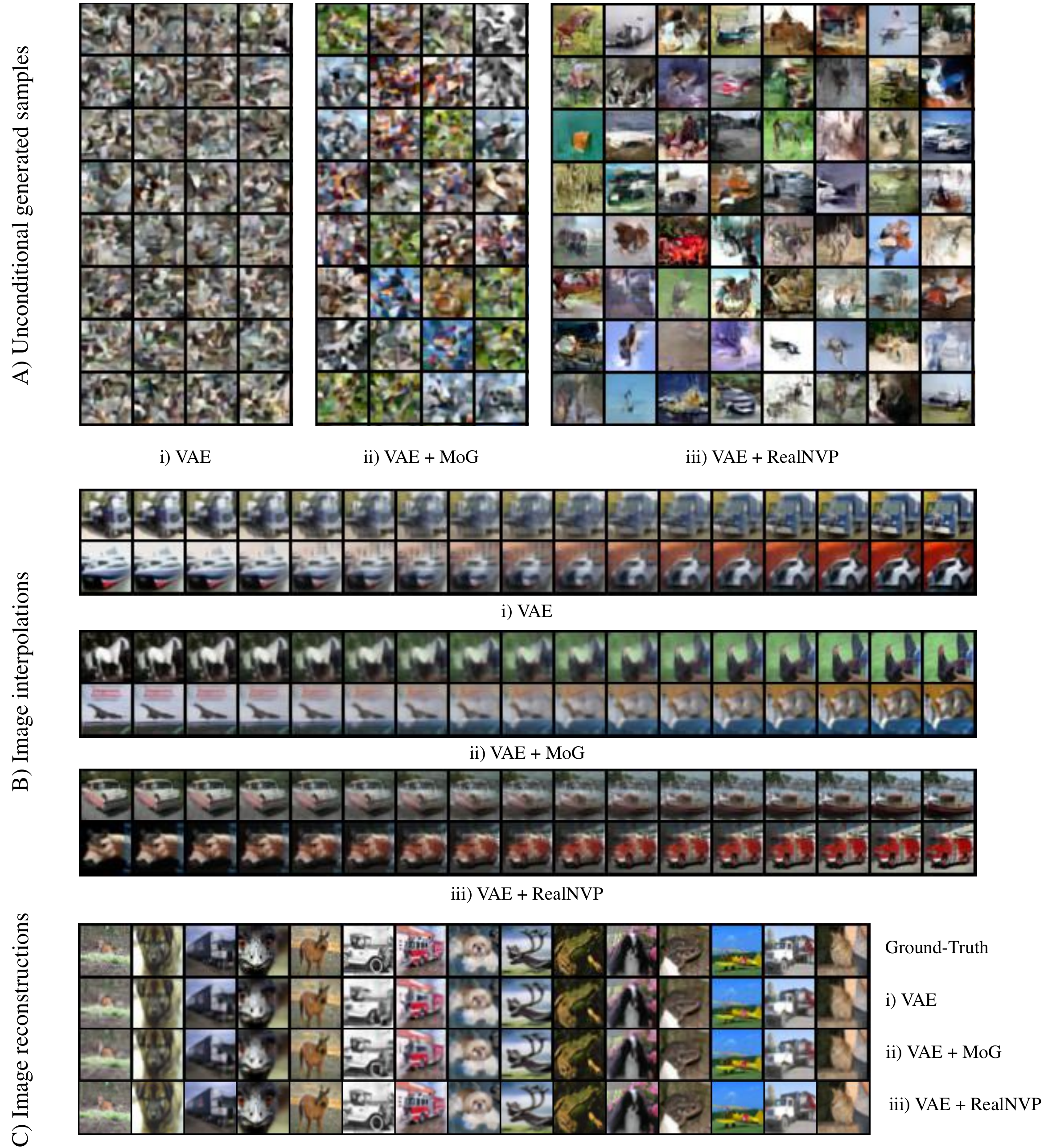}
    \end{subfigure}}
    \caption[VAE prior comparison]{Qualitatively results on CIFAR-10 for VAEs with various latent prior distributions. The depicted results show A) unconditional generative samples B) image interpolations and C) image reconstructions.}
    \label{fig:prior_results}
\end{figure}

\paragraph{Quantitative Analysis} Table \ref{tab:prior_nll} represents the quantitative results of density estimation on the natural image dataset CIFAR-10. We measure performance by estimating the log-likelihood with 512 importance weighted samples on the test set. To begin with, the use of a mixture of Gaussians, a learned distribution, performs better that the VAE with fix prior, with a negligible cost on additional trainable parameters, which do not allocate extra time in sample generation. However, the adaption of RealNVP as a bijective network, improves significantly the likelihood score. Interestingly, both models with learned priors perform better than the fixed one, by reaching better reconstruction loss with the cost of allocating more regularization cost. This behavior exactly reflects the relationship between variational posterior and the prior; when a fix unimodal Gaussian is used, it pulls the variational ones to its pick. As a result, all the variational posteriors are close to the mean of the prior, resulting into a smaller KL diverge but blurry reconstructions. On the contrary, the learned multi-modal priors allow the posteriors to move more freely and are able to adjust to the variational family (see figure \ref{fig:prior}). This distribution mismatch is the greatest when the RealNVP prior is used, indicating that the prior allows the encoder to form more and more complex distributions, by trading of regularization loss for better reconstructions.

\paragraph{Qualitative Analysis} The difference in performance becomes more obvious by looking at the qualitatively results, presented in Figure \ref{fig:prior_results}. Looking at the generative samples, we see a clear improvement when moving to richer prior. The plain VAE with standard normal distribution generates random patters while the model with a mixture of Gaussian prior enhance them with colors. However, only VAE with RealNVP showcases generations that manage to display a global context. To examine the richness of the learnt space, we also perform interpolation on the latent codes between two ground-truth images. Ideally, the internal generations should result into meaningful modifications of the provided samples. In the case of VAE, we see that the inner generations produce blurry samples, indicating the unexpressivety of the latent codes. The generations are getting better with the MoG prior, but the most impressive results are coming from the framework with RealNVP. We can see that the samples are more influenced by their closest ground-truth image, while adapting more and more core characteristics when moving closer to the other. For example, on the second row of B:iii, the original brown horse first turns into red, adapting the color of the firetruck, before it smoothly result into it. Lastly, on image reconstructions we again witness the same patent, when the richer prior does result into better reconstructions, confirming the quantitatively results presented on table \ref{tab:prior_nll}. Specifically, the VAE with the bijective prior showcases an excellent performance on the natural image reconstruction task, which is contrary to the performance of the other two approaches. This is associated with the effectiveness of the powerful, invertible, data-driven prior (in our case, the RealNVP) and its ability to boost the performance significantly with negligible sacrifice on generation speed, and none on inference.

\begin{table*}[t]
\caption[VAE with different priors for CIFAR-10]{Generative modelling performance on CIFAR-10 obtained from different priors in bits per dimension. The generation time is based on $100$ runs on a singe GeForce GTX 1080 Ti GPU for 1 image (and 100 images).}
\centering
\scalebox{.79}{
\begin{tabular}{lccccccc}
\multirow{2}{*}{\textbf{Model}} & \multirow{2}{*}{\textit{\begin{tabular}[c]{@{}c@{}}nll\\ (bits/dim)\end{tabular}}} & \multicolumn{2}{c}{\textit{reconstruction loss}} & \multicolumn{2}{c}{\textit{regularization loss}} & \multirow{2}{*}{\#params} & \multirow{2}{*}{\textit{generation time} (sec)}\\
&  & \multicolumn{2}{c}{$\text{RE}_{x}$} & \multicolumn{2}{c}{$\text{KL}_{z}$} \\ \hline
VAE & 3.88 & \multicolumn{2}{c}{\textit{6675}} & \multicolumn{2}{c}{\textit{1596}} & 20M & 0.01 (0.07) \\ 
VAE + MoG & 3.83 & \multicolumn{2}{c}{\textit{6512}} & \multicolumn{2}{c}{\textit{1653}} & 22M & 0.01 (0.07) \\ 
VAE + RealNVP & \textbf{3.51} & \multicolumn{2}{c}{\textit{5540}} & \multicolumn{2}{c}{\textit{1966}} & 32M & 0.07 (0.14)\\
\bottomrule
\end{tabular}}
\label{tab:prior_nll}
\end{table*}


\subsection{Neural Network Architecture}
\label{sec:nn}

The choice of the NN architecture is crucial for the performance and the scalability of the overall framework, and usually architectures that showcased great performance in discriminate tasks (i.e. classification) are used in generative modelling tasks as well. However, the internal representations that the networks have to discover are fundamentally different, and little attention has been given into designing a NN specifically for an auto-encoder setting. For example, in classification tasks the network extracts specific representation of a particular object, in contrast with the generative models, where we aim for discovering the semantic structure of the data. Thus, as we argue that we can benefit from a carefully designed architecture, in this section we present our approach.

For building blocks of the network, we employed densely connected convolutional networks instead of residual ones. The motivation for this choice is that since DenseNets encourage feature reuse, it will help preserve visual information from the very first layer effectively, while requiring less trainable parameters. Thus, the network could discover easier generic graphical features and local pixel correlations. The concatenation of the filters will also alleviate the vanishing-gradient problem and allow us to build deep architectures. Additionally, exponential linear units (ELUs) are used everywhere as activation functions. In contrast to ReLUs, ELUs have negative values which allows them to push mean unit activations closer to zero, which speeds up the learning process. This is due to a reduced \textit{bias shift effect}; bias that is introduced to the units from those of the previous layer which have a non-zero mean activation.

Typically, every convolution operation precedes a batch normalization layer, as they empirically exhibit a boost in performance in discriminate tasks. However, their performance is known to degrade for small batch sizes, as the variance of the activation noise that they contribute is inversely proportional to the number of data that is processed. This noise injection, in combination with their intensive memory demands, can be critical drawbacks when we process image data, especially high-dimensional ones. We instead use weight normalization, where even though it separates the weight vector from its direction just like batch normalization, they do not make use of the variance. This allows them to get the desired output even in small mini-batches, while allocating a small proportion of memory. We empirically find out that indeed using weight normalization reduces the overfitting of the model. In addition, we used a data-dependant initialization of the model parameters, by sampling the first batch of the training set. This will allow the parameters to be adjusted by the output of the previous layers, taking into account and thus resulting into a faster learning process.

An important element of the auto-encoding scheme is the process of feature downscaling and upscaling. Despite its success in classification tasks, pooling is a fixed operation that replacing it with a stride-convolution layer can also be seen as generalization, as the scaling process is now learned. This will increase the models' expressibility with the cost of adding a negligible amount of learning parameters. For the upscaling operation, even though various methods have been proposed \citep{shi2016realtime}, we found out that the plain transposed convolution generalised better than the others, while requiring far less trainable parameters. Finally, inspired from the recent advantages on super-resolution neural network architectures, we used \textit{channel-wise attention} blocks (CA) at the end of every DenseNet block \citep{zhang2018image}. The CA blocks will help the network to focus on more informative features, by exploiting the inter-dependencies among feature channels. Thus, it performs feature recalibration in a global way, where the per-channel summary statistics are calculated and then used to selectively emphasise informative feature-maps as well as suppress useless ones (e.g. redundant feature-maps). This is done through a global average pooling, that squeezes global spatial information into a channel statistical descriptor, followed by a gating mechanism, where it learns nonlinear interactions between the input channels. 

\begin{figure}[h]
    \centering
    \scalebox{.25}{
    \begin{subfigure}{1.\textwidth}
      \centering
      \includegraphics[width=1.\linewidth]{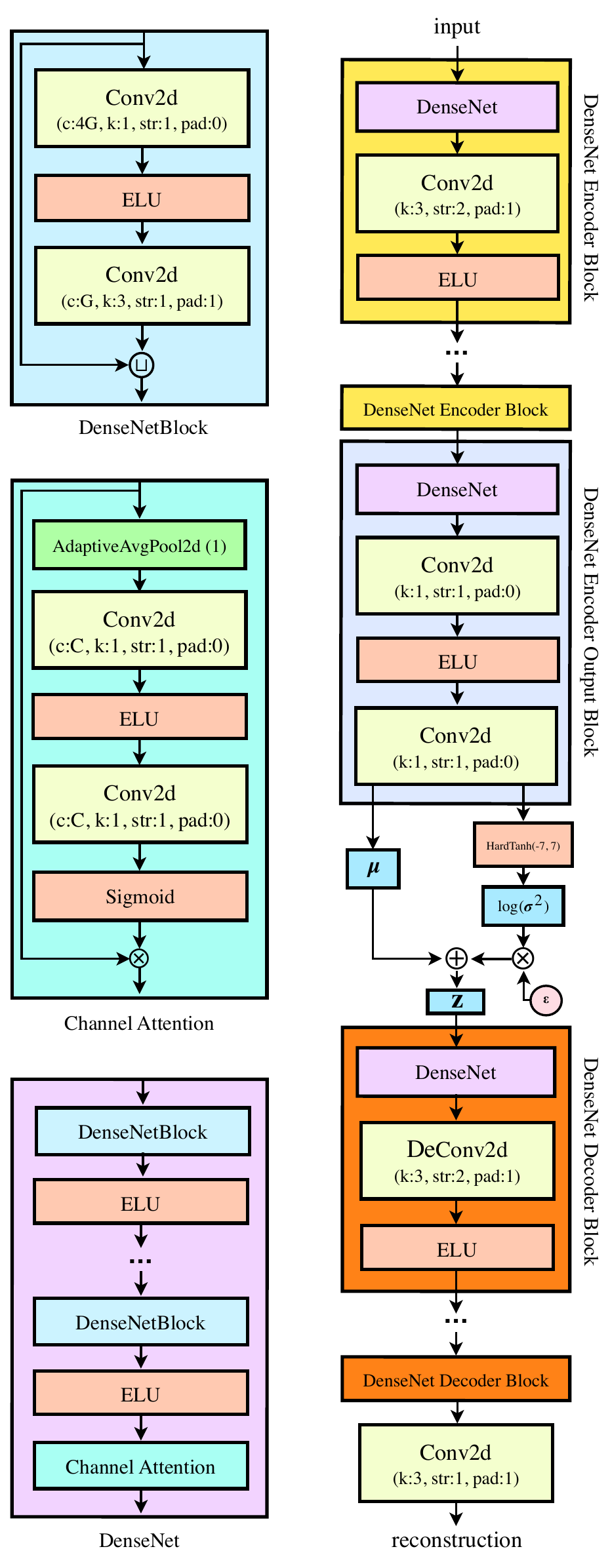}
    \end{subfigure}}
    \caption[Architecture of the autoencoder]{Architecture of our autoencoder. On the right, there are some basic buildings block of the network. The notation as 'G' on the Conv2D channels indicate the growth rate of the densely connected network. The $\epsilon$ indicates a random variable drawn from a standard Gaussian, which helps us to make use of the \textit{reparametrization} trick. Until $\mathbf{z}$, we refer to this architecture as \textit{Encoder NN} and thereafter as \textit{Decoder NN}. The former and the later form the \textit{building blocks} to every model that we train and evaluate.}
    \label{fig:nn_architecture}
\end{figure}

The core building blocks and the network of an auto-encoding network are illustrated in Figure \ref{fig:nn_architecture}.

\subsection{Derivation of the lower bound}
\label{sec:elbo_derivation}

Expanding the lower bound from (\ref{eq:our_elbo}), from the first part we will have

\begin{gather*}
    \E\mathord{_{q(\mathbf{w})}}
    \Big[
    \log p(\mathbf{x}, \mathbf{w})
     \Big]
    =
    \E\mathord{_{q(\mathbf{w})}}
    \Big[
    \log 
    p(\mathbf{x}|\mathbf{y}, \mathbf{z}) \,
    p(\mathbf{z}|\mathbf{y}, \mathbf{u}) \,
    p(\mathbf{y}|\mathbf{u}) \,
    p(\mathbf{u})
    \Big]
    =
    \E\mathord{_{q(\mathbf{z}|\mathbf{y}, \mathbf{x}) q(\mathbf{y}|\mathbf{x})}}
    \Big[ \log p(\mathbf{x}|\mathbf{y}, \mathbf{z}) \Big]
    +\\+
    \E\mathord{_{q(\mathbf{z}|\mathbf{y}, \mathbf{x}) \,
    q(\mathbf{u}|\mathbf{y}) \, q(\mathbf{y}|\mathbf{x})}}
    \Big[ 
    \log
    p(\mathbf{z}|\mathbf{y}, \mathbf{u})
    \Big]
    +
    \E\mathord{_{q(\mathbf{u}|\mathbf{y}) q(\mathbf{y}|\mathbf{x})}}
    \Big[ \log p(\mathbf{y}|\mathbf{u}) \Big] \,
    + 
    \E\mathord{_{q(\mathbf{u}|\mathbf{y}) q(\mathbf{y}|\mathbf{x})}}
    \Big[ \log p(\mathbf{u}) \Big],
\end{gather*}

and for the second

\begin{align*}
    \E\mathord{_{q(\mathbf{w})}}
    \Big[
    \log q(\mathbf{w})
    \Big]
    =
    \E\mathord{_{q(\mathbf{z}|\mathbf{y}, \mathbf{x})}}
    \Big[ 
    \log
    q(\mathbf{z}|\mathbf{y}, \mathbf{x})
    \Big]
    +
    \E\mathord{_{q(\mathbf{y}|\mathbf{x})}}
    \Big[ \log q(\mathbf{y}|\mathbf{x}) \Big] \,
    +
    \E\mathord{_{q(\mathbf{u}|\mathbf{y}) \, q(\mathbf{y}|\mathbf{x})}}
    \Big[ \log q(\mathbf{u}|\mathbf{y}) \Big].
\end{align*}

Interestingly, plugging the above terms back to (\ref{eq:our_elbo}) and rearranging them, we will have

\begin{gather*}
    \mathcal{L}(\mathbf{x})
    \overset{(\ref{eq:our_elbo})}{=}
    \underbrace{
    \E\mathord{_{q(\mathbf{z}|\mathbf{y}, \mathbf{x}) q(\mathbf{y}|\mathbf{x})}}
    \Big[ \log p(\mathbf{x}|\mathbf{y}, \mathbf{z}) \Big] \,
    +
    \E\mathord{_{q(\mathbf{z}|\mathbf{y}, \mathbf{x}) \,
    q(\mathbf{u}|\mathbf{y}) \, q(\mathbf{y}|\mathbf{x})}}
    \Big[ 
    \log
    p(\mathbf{z}|\mathbf{y}, \mathbf{u})
    \Big] - \E\mathord{_{q(\mathbf{z}|\mathbf{y}, \mathbf{x})}}
    \Big[ 
    \log
    q(\mathbf{z}|\mathbf{y}, \mathbf{x})
    \Big]
    }_{\boldsymbol{A}}
    +
    \\+
    \underbrace{
    \E\mathord{_{q(\mathbf{u}|\mathbf{y}) q(\mathbf{y}|\mathbf{x})}}
    \Big[ \log p(\mathbf{y}|\mathbf{u}) \Big] \,
    + 
    \E\mathord{_{q(\mathbf{u}|\mathbf{y}) q(\mathbf{y}|\mathbf{x})}}
    \Big[ \log p(\mathbf{u}) \Big] \,
    -
    \E\mathord{_{q(\mathbf{y}|\mathbf{x})}}
    \Big[ \log q(\mathbf{y}|\mathbf{x}) \Big] \,
    -
    \E\mathord{_{q(\mathbf{u}|\mathbf{y}) q(\mathbf{y}|\mathbf{x})}}
    \Big[ \log q(\mathbf{u}|\mathbf{y}) \Big]
    }_{\boldsymbol{B}}.
\end{gather*}

Working with term $\boldsymbol{B}$, one can see that

\begin{gather*}
    \boldsymbol{B}
    = 
    \E\mathord{_{q(\mathbf{u}|\mathbf{y}) q(\mathbf{y}|\mathbf{x})}}
    \Big[
    \log 
    \frac{p(\mathbf{y}|\mathbf{u}) p(\mathbf{u})}{q(\mathbf{u}|\mathbf{y}) q(\mathbf{y}|\mathbf{x})}
    \Big],
\end{gather*}

which denotes a (hidden) lower bound on of the marginal $\log p(\mathbf{y})$
with variational posterior $q(\mathbf{u}|\mathbf{y}) q(\mathbf{y}|\mathbf{x})$.

Thus, the resulted lower bound of the marginal likelihood of $\mathbf{x}$ would be

\begin{align*}
    \mathcal{L}(\mathbf{x})
    &=
    \E\mathord{_{q(\mathbf{z}| \mathbf{x}, \mathbf{y}) \, q(\mathbf{y}|\mathbf{x})}} \log 
    p_{\theta}(\mathbf{x}| \mathbf{y}, \mathbf{z}) \,  
    - \KL({q(\mathbf{z}| \mathbf{x}, \mathbf{y}) || p(\mathbf{z}| \mathbf{y}, \mathbf{u}}))
    + \E\mathord{_{q(\mathbf{u})}} \log 
    p_{\theta}(\mathbf{y}| \mathbf{u})
    -
    \KL({q(\mathbf{u}| \mathbf{y}})||p(\mathbf{u}) )
    \\&=
    \E\mathord{_{q(\mathbf{z}| \mathbf{x}, \mathbf{y}) \, q(\mathbf{y}|\mathbf{x})}} \log 
    p_{\theta}(\mathbf{x}| \mathbf{y}, \mathbf{z}) \,  
    - \KL({q(\mathbf{z}| \mathbf{x}, \mathbf{y}) || p(\mathbf{z}| \mathbf{y}, \mathbf{u}}))
    +
    \E\mathord{_{q(\mathbf{u}|\mathbf{y}) q(\mathbf{y}|\mathbf{x})}}
    \Big[
    \log 
    \frac{p(\mathbf{y}|\mathbf{u}) p(\mathbf{u})}{q(\mathbf{u}|\mathbf{y}) q(\mathbf{y}|\mathbf{x})}
    \Big].
\end{align*}




\subsection{Datasets}
\label{sec:datasets}

\textbf{CIFAR-10}\quad The CIFAR-10 dataset is a well-known image benchmark data containing 60.000 training examples and 10.000 validation examples. From the training data, we put aside $15\%$ randomly selected images as the test set. We augment the training data by using random horizontal flips and random affine transformations and normalize the data uniformly in the range (0, 1).

\textbf{Imagenette64}\quad Imagenette64\footnote{https://github.com/fastai/imagenette} is a subset of 10 classes from the downscaled Imagenet dataset. We downscaled the dataset to $64$px $\times$ $64$px images. Similarly to CIFAR-10, e put aside $15\%$ randomly selected training images as the test set. We used the same data augmentation as in CIFAR-10

\textbf{CelebA}\quad The Large-scale CelebFaces Attributes (CelebA) Dataset consists of $202.599$ images of celebrities.
We cropped original images on the 40 vertical and 15 horizontal component of the top left corner of the crop box, which height and width were cropped to 148. Besides the uniform normalization of the image, no other augmentation was applied.


\subsection{Self-Supervised VAE - Sketch Reconstructions}
\label{sec:recon_sketch}

\begin{figure}[h]
\centering
\scalebox{.99}{
\begin{subfigure}{1.\textwidth}
  \centering
  \includegraphics[width=1.\textwidth]{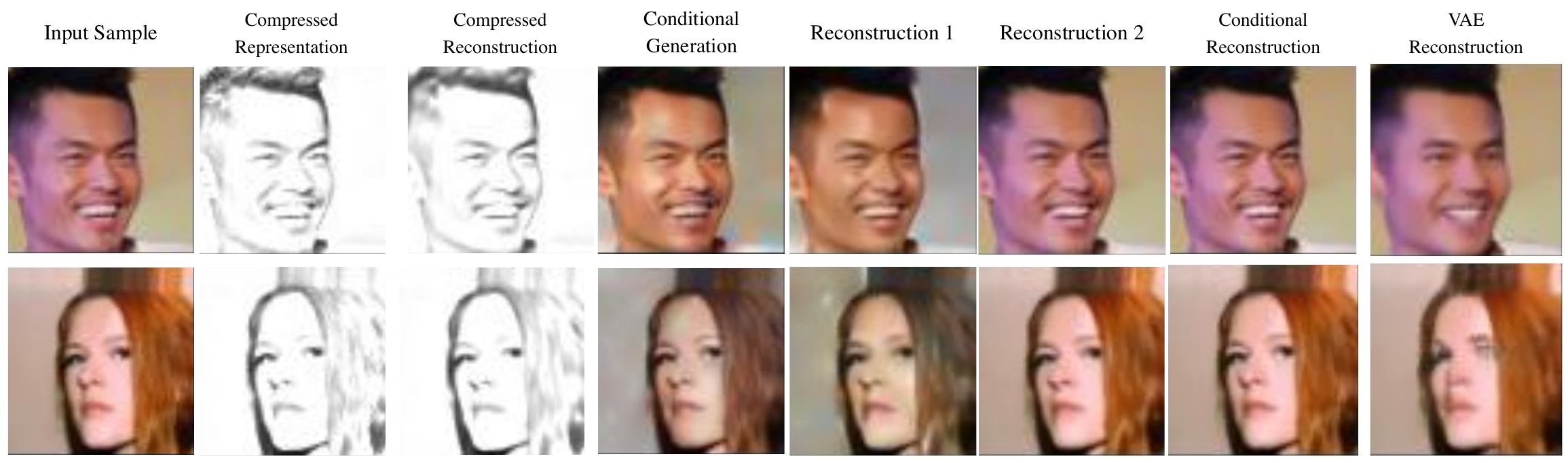}
\end{subfigure}}
\caption{Qualitative results illustrating all the reconstruction techniques on CelebA for selfVAE-sketch.}
\label{fig:recon_sketch}
\end{figure}

Given the astonishing performance on visual tasks, CNNs are commonly thought to recognise objects by learning increasingly complex representations of object shapes. However, \cite{geirhos2018imagenettrained} showed that where humans see shapes, CNNs are strongly biased towards recognising textures. Furthermore, architectures that learn shape-based representations come with several unexpected emergent benefits such as previously unseen robustness towards a wide range of image distortions. This acted as a motivation to employ the framework of self-supervised auto-encoder with a representation that captures the shape of the object. Thus, the first part will model the outlines of the given object while the second will be responsible for its texture.

We can retrieve a shape-based representation of an image by detecting its edges. Edges appear when there is a sharp change in brightness and it usually corresponds to the boundaries of an object. There are many different techniques for computing the edges, like using filters that extract the gradient of the image (Sobel kernels, Laplacian of Gaussian, etc.). In this experiment, we will use the method proposed in \cite{GastalOliveira2011DomainTransform}, as it is a fast, high-quality edge-preserving filtering of images. Specifically, in order to obtain a pencil \textit{sketch} (that is, a black-and-white drawing) of the input image, we will make use of two image-blending techniques, known as \textit{dodging} and \textit{burning}. The former lightens an image, whereas the latter darkens it. A sketch transformation can be obtained by using dodge to blend the grayscale image with its blurred inverse. In this way, we produce high-quality edge-preserving filtering of the image efficiently performed in linear time.

\paragraph{Qualitative Analysis} Similar to the case when we used a downscaled transformation of the input image as conditional representation, the selfVAE framework here also allows for different ways to reconstruct an image. The qualitative results when we employ a sketch representation are visible in Figure \ref{fig:recon_sketch}. To start with, we see that even though because of the sketch representation we lose the texture of the image, we preserve not only the global information but also high-level details that characterize each person. In this way, we let the generative model emphasise on these specifics at its first step, which through the latent variable $\mathbf{u}$, manages to reconstruct with tremendous accuracy. This can be confirmed by visually comparing the images of the original compressed images (CM) with those of the reconstructed ones. These great results hold also for the conditional reconstruction, where the original sample (OG) is synthesised from conditioning on the original sketch representation (CM). Furthermore, we gain further interpretation of the model through the reconstructions that use only the latent variable $\mathbf{u}$ (RS1) and both latent variables (RS2). Given that both methods infer the sketch image, the end results still preserve all detail concepts of the human portrait. However, they are different in terms of the texture (colour) of the image, which is modelled through $\mathbf{z}$. One the one hand, in the first case its value is inferred through the sketch, and thus it produces the most probable values. That is why we see on ground-truth images that incorporate unusual lighting, it outputs a more natural outcome. On the other hand, in the second case, the latent codes of $\mathbf{z}$ are computed given the original sample. And from the results, we can see that, indeed, all the information about the texture of the image can be compressed into the latent $\mathbf{z}$, as these reconstructions are identical with the input ones. Additionally, the compressed generation (CG) process is alike to this of RS1 as they both infer the texture but different as it uses the ground-truth sketch representation. It is worth to note that the end result in both cases share the same quality, which is another indication that the reconstructions of the sketch images are excellent.

\subsection{Interpolation through latent spaces and conditional generations}
\label{sec:latent_space}

\begin{figure}[h]
\centering
\scalebox{.99}{
\begin{subfigure}{1.\textwidth}
  \centering
  \includegraphics[width=1.\textwidth]{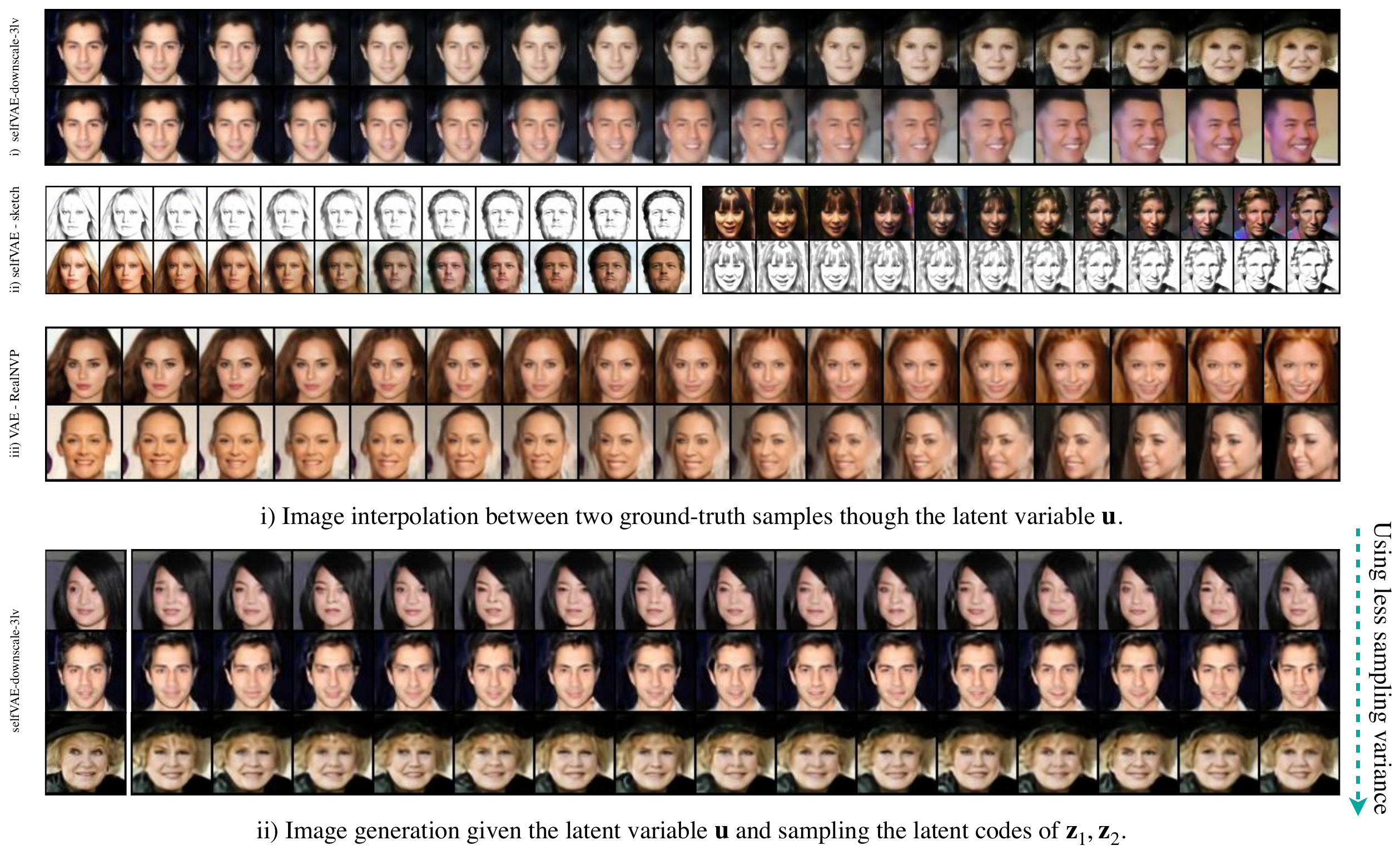}
\end{subfigure}}
\caption{Latent space interpolations and conditional generations of the selfVAEs.}
\label{fig:latent}
\end{figure}

In Figure \ref{fig:latent} we visualise i) interpolations through two ground-truth images through the latent code $\mathbf{u}$ and ii) image reconstructions, where we keep the latent code $\mathbf{u}$ but varying all the others ($\mathbf{z}_1$ and $\mathbf{z}_2$) of the 3-leveled selfVAE architecture. Just like the previous cases, in the first case, we see that the model incorporated a rich latent space $\mathbf{U}$, which is responsible for the generation and construction of the global structure of the image. Moving from one latent code $\mathbf{u}$ of a given image to another, we obtain meaningful modifications of the image that result in images that share characteristics from both of them. However, in the latter case, we see that we can alter only high-level features of the image when we keep the values of $\mathbf{u}$ but vary the others; $\mathbf{z}_1$ and $\mathbf{z}_2$. Interestingly, we see that given that ground-truth image that is illustrated on the very left, we can sample different expressions and characteristics of the same person, as the latent variable $\mathbf{u}$ is kept constant.

\subsection{Additional Results}
\label{sec:extra_results}

\begin{figure}[t]
\centering
\scalebox{.99}{
\begin{subfigure}{1.\textwidth}
  \centering
  \includegraphics[width=1.\linewidth]{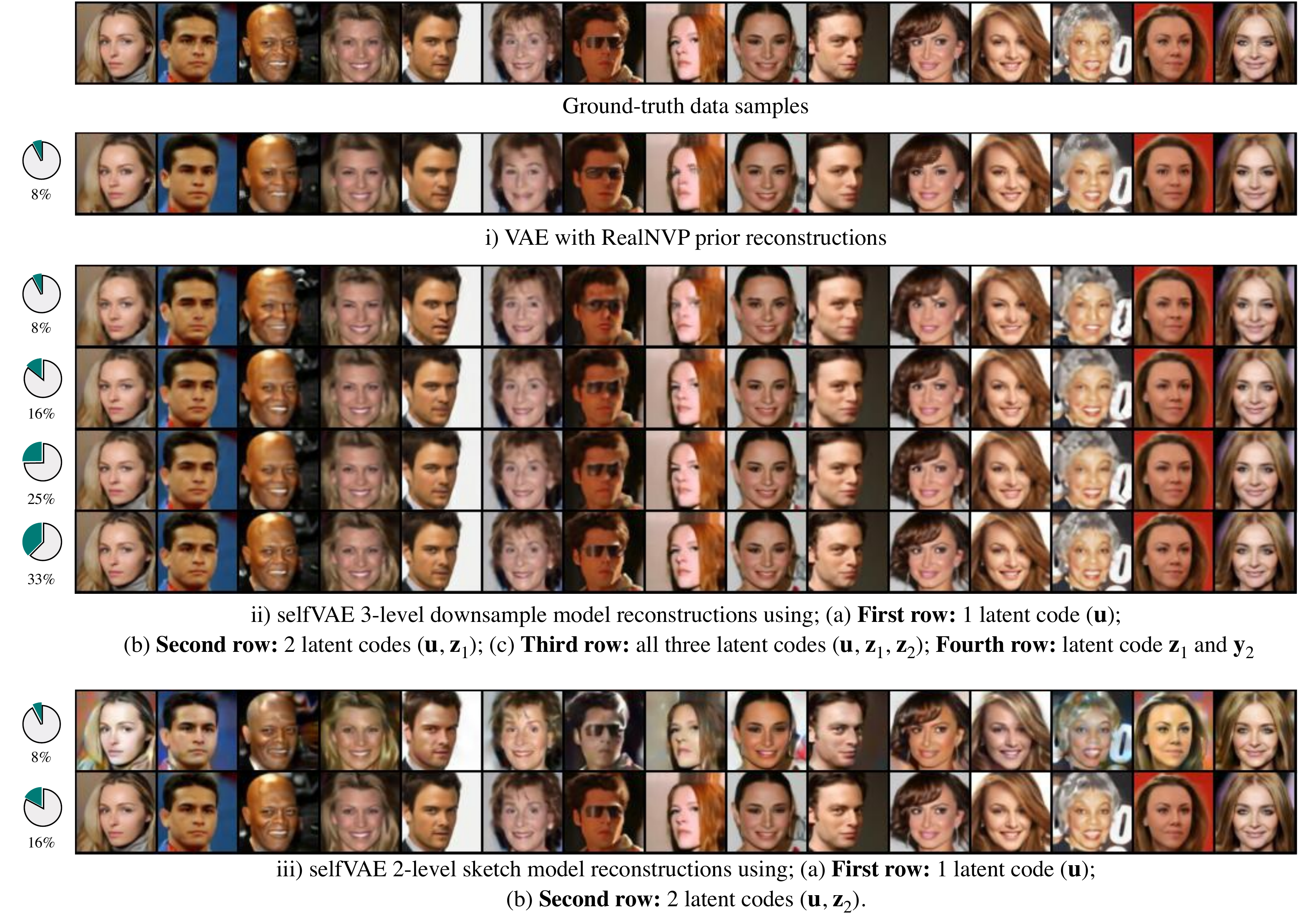}
\end{subfigure}}
\vskip -3mm
\caption{Comparison on image reconstructions with different amount of sent information.}
\label{fig:celeba_comp_recon_1}
\end{figure}

Additional results of reconstructions for CelebA are shown in Figure \ref{fig:celeba_comp_recon_1}.

\end{document}